\PassOptionsToPackage{table,dvipsnames}{xcolor} 
\documentclass[10pt,twocolumn,letterpaper]{article}

 \usepackage{iccv}              
\newcommand{\tablestyle}[2]{\setlength{\tabcolsep}{#1}\renewcommand{\arraystretch}{#2}\centering\footnotesize}
\usepackage{makecell}
\usepackage{multicol}
\usepackage{multirow}
\usepackage{xcolor} 
\usepackage{colortbl} 
\definecolor{gr}{gray}{0.95} 
\newcommand{\gr}{\rowcolor{gr}} 
\definecolor{convcolor}{HTML}{412F8A}   
\definecolor{vitcolor}{HTML}{fc8e62}    

\newcommand{\convcolor}[1]{\textcolor{convcolor}{#1}}
\newcommand{\vitcolor}[1]{\textcolor{vitcolor}{#1}}

\newcommand{\vb}{\vitcolor{$\mathbf{\circ}$\,}}  
\newcommand{\cb}{\convcolor{$\bullet$\,}}        

\definecolor{iccvblue}{rgb}{0.21,0.49,0.74}
\usepackage[pagebackref,breaklinks,colorlinks,allcolors=iccvblue]{hyperref}
\usepackage[accsupp]{axessibility}

\newcommand{\whitefootnote}[1]{%
  \begingroup
  \renewcommand\thefootnote{\textcolor{white}{\arabic{footnote}}}%
  \hypersetup{hidelinks}
  \footnote{\textcolor{black}{#1}}
  \endgroup
}
\title{INTER: Mitigating Hallucination in Large Vision-Language Models by Interaction Guidance Sampling}

\author{%
  Xin Dong$^{*,1,2}$, Shichao Dong$^{*,2}$, Jin Wang$^{*,3}$, Jing Huang$^{\dagger,1,4}$, Li Zhou$^{2}$, Zenghui Sun$^{2}$,\\  Lihua Jing$^{1,4}$, Jinsong Lan$^{2}$, Xiaoyong Zhu$^{2}$, Bo Zheng$^{\dagger,2}$\\
  \textsuperscript{1}University of Chinese Academy of Sciences 
       \textsuperscript{2}Taobao $\&$ Tmall Group of Alibaba \\
        \textsuperscript{3}The University of Hong Kong 
        \textsuperscript{4}Institute of Information Engineering, Chinese Academy of Sciences\\
           \tt\small \{dongxin,huangjing,jinglihua\}@iie.ac.cn,
        \{dongshichao1996\}@gmail.com,\{wj0529\}@connect.hku.hk,\\
      \tt\small  \{pengye.zl,zenghui.szh,jinsonglan.ljs\}@taobao.com,
      \{xiaoyong.z,bozheng\}@alibaba-inc.com
}

\begin{document}

\maketitle
\begin{abstract}
Hallucinations in large vision-language models (LVLMs) pose significant challenges for real-world applications, as LVLMs may generate responses that appear plausible yet remain inconsistent with the associated visual content.
This issue rarely occurs in human cognition. 
We argue that this discrepancy arises from humans' ability to effectively leverage multimodal interaction information in data samples. 
Specifically, humans typically first gather multimodal information, analyze the interactions across modalities for understanding, and then express their understanding through language.
Motivated by this observation, we conduct extensive experiments on popular LVLMs and obtained insights that surprisingly reveal human-like, though less pronounced, cognitive behavior of LVLMs on multimodal samples. 
Building on these findings, we further propose \textbf{INTER}: \textbf{Inter}action Guidance Sampling, a novel training-free algorithm that mitigate hallucinations without requiring additional data. 
Specifically, INTER explicitly guides LVLMs to effectively reapply their understanding of multimodal interaction information when generating responses, thereby reducing potential hallucinations. 
On six benchmarks including VQA and image captioning tasks, INTER achieves an average improvement of up to 3.4\% on five LVLMs compared to the state-of-the-art decoding strategy. 
The codes are released on \href{https://github.com/xxxxx313/INTER}{Github}. 
\end{abstract}   
\whitefootnote{$^*$ Equal Contribution. } \\ 
\whitefootnote{$^\dagger$ Corresponding authors.}
\section{Introduction}
\label{sec:intro}
\begin{figure}[]
    \centering
    \includegraphics[width=1.0\linewidth, trim = 0 0 0 67,clip]{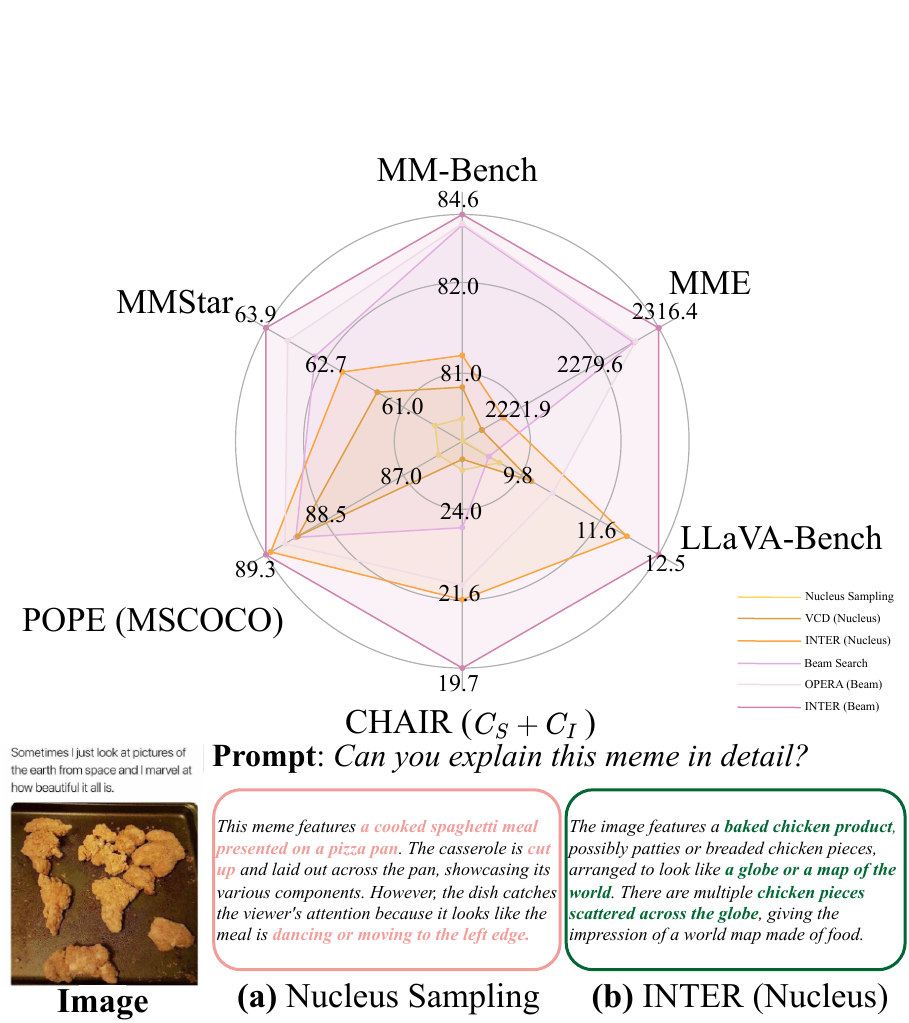}
    \caption{
    \textbf{Comparisons with existing decoding strategies on the state-of-the-art LVLM InternVL2.5-MPO~\cite{chen2024internvl}.} Our approach INTER achieved optimal results across six benchmarks. Besides, the detailed output of INTER in a case of complex inputs is presented. The hallucinated words are highlighted in red. }
    \label{fig:intro}
\end{figure}
Large Vision-Language Models (LVLMs) have demonstrated remarkable versatility across a wide range of tasks, from image captioning to complex reasoning~\cite{bai2023qwen, li2022mplug, ye2024mplug, chen2024internvl, liu2023improvedllava, instructblip}. 
These models blend visual and textual information, enhancing our ability to interpret and interact with the world. 
However, LVLMs experience hallucinations that hinder their applications, which means that responses are not aligned with the given input~\cite{huang2024opera,leng2023mitigating,jiang2024hallucination,chen2024halc,hsun2023aligning,hwang2023llm,hchen2023mitigating}.

Previous methods made attempts to address this issue with additional training and fine-grained data~\cite{liu2023mitigating, yu2024hallucidoctor,zhai2023halle,jain2024vcoder,hsun2023aligning,yue2024less}, but such approaches often demand substantial human effort and computational resources. 
In parallel, other methods were proposed to explore efficient training-free methods to mitigate hallucinations in LVLMs~\cite{leng2023mitigating, huang2024opera, favero2024multi, chen2024halc,park2024convis,zou2024look,an2024agla,qu2024look}. 
These methods aimed to enhance the model's focus on the input \textit{visual} or \textit{textual} information by adjusting the logit distribution during the autoregressive generation process, thereby reinforcing the link between the output and \textit{uni-modal} information.  

Despite these efforts, the role of \textit{multimodal interactions} information in shaping LVLMs' hallucinated responses remains largely underexplored, even though it naturally plays a critical role in human cognition for accurate reasoning.
Specifically, when presented with a textual prompt and an associated image, humans typically first gather multimodal information, analyze the interactions between modalities to form a conceptual understanding, and then provide textual responses based on this understanding.

Motivated by this intuition, we propose to first investigate whether LVLMs exhibit similar cognitive behavior when processing multimodal data, aiming to uncover the potential causes behind their hallucination issues.
To this end, we designed several metrics based on the Harsanyi dividend~\cite{harsanyi1982simplified} to explicitly quantify the influence of image-text multimodal interactions in LVLMs' responses.
The Harsanyi dividend was originally proposed in game theory, which measures the interactions among different players. 
Associated with the Shapley value~\cite{shapley1953value}, the Harsanyi dividend theoretically satisfies the \textit{efficiency}, \textit{linearity}, \textit{dummy}, \textit{symmetry} axioms, which further enhances its trustworthiness on the analyses of LVLMs ~\cite{ren2021can,ren2023defining,wang2024diagnosing}. 

In this way, we derive the following key insights through extensive analysis:
\begin{itemize}[noitemsep,topsep=0pt,left=0pt]
\item \textit{Insight 1}: \textbf{LVLMs implicitly capture multimodal interactions from input samples and leverage such interactions for decision-making to some extent.}
\item \textit{Insight 2}: \textbf{LVLMs exhibit a tendency to primarily apply their understanding of multimodal interactions to the generation of a few key tokens, rather than uniformly across all response tokens. }
\item \textit{Insight 3}: \textbf{The understanding of multimodal interactions in LVLMs positively influences the quality of generated responses, with stronger interactions leading to greater accuracy. }
\end{itemize}
To our surprise, the above findings suggest that LVLMs possess a human-like—albeit less pronounced—understanding of multimodal interactions, which inspired us to design \textit{efficient} methods for reducing hallucinated responses of LVLMs. 

To mitigate the hallucinations of LVLMs, we propose a simple yet effective method named \textbf{INTER}: \textbf{Inter}action Guidance Sampling, which aims to reapply their captured multimodal interaction understanding (\textit{cf}. \textit{Insight 1}) more accurately and effectively in LVLMs' reasoning process. 
Specifically, we first design a variance-based filtering module termed as the Interactive Guided Locator in INTER to automatically detect key tokens that significantly contribute to the accuracy of responses (\textit{cf}. \textit{Insight 2}). 
After that, we design the Interaction Probability Modifier in INTER which guides the sampling of these key tokens to rely more on multimodal interactions (\textit{cf}. \textit{Insight 3}). 
In this way, LVLMs can reduce potential hallucinations in their responses, improving their overall performance. 

As shown in \cref{fig:intro}, our INTER achieved better performance on the state-of-the-art LVLM InternVL2.5-MPO (8B)~\cite{chen2024internvl} across six widely-used benchmarks \cite{li2023evaluating,liu2023mmbench,chen2024we,fu2023mme,rohrbach2018object,llava_bench_in_the_wild}. 
In experiments, we also verified the effectiveness of INTER on other popular LVLMs \cite{instructblip,liu2023improvedllava,ye2024mplug,bai2023qwen,chen2024far}. 
Specifically, 
INTER boosted the performance of 
state-of-the-art decoding strategies 
by an average of 
 4.1\% and 2.6\% on CHAIR~\cite{rohrbach2018object} and MME~\cite{fu2023mme} benchmark, respectively. 

Our contributions can be summarized as follows:  
\begin{itemize}[noitemsep,topsep=0pt,left=0pt]
    \item From a novel game-theoretic view, we propose to investigate the roles of multimodal interactions in shaping LVLMs’ hallucinated responses. Through extensive analyses, we obtain several new insights accordingly.  
    \item We propose Interaction Guidance Sampling (INTER), a novel plug-and-play sampling rectification method for eliminating hallucinations, that accurately and effectively guides LVLMs to explicitly reapply their understanding of multimodal interactions in responses. 
    \item Extensive experiments demonstrate that INTER successfully improved the performance of various LVLMs upon six benchmarks without requiring additional training, outperforming state-of-the-art methods by a large margin. 
\end{itemize}

\section{Related Work}
\label{sec:related}

\subsection{Large Vision-Language Models (LVLMs)}
The rapid advancement of Large Vision-Language Models (LVLMs) has become a pivotal area in artificial intelligence research. 
These models aim to generate contextually grounded responses through multimodal understanding. 
Modern architectures typically adapt existing Large Language Models (LLMs) as text decoder for generation.  
Prominent examples include LLaVA-v1.5~\cite{liu2023improvedllava} and InstructBLIP~\cite{instructblip}, both built on Vicuna 7B~\cite{chiang2023vicuna}, as well as Qwen-VL~\cite{bai2023qwen} and mPLUG-Owl2~\cite{ye2024mplug}, which utilize Qwen 7B~\cite{bai2023qwenlm} and LLaMA 2 7B~\cite{touvron2023llama2} respectively. Scalability efforts are exemplified by InternVL2~\cite{chen2024internvl}, which explores parameter configurations from 1B to 108B. 
Despite these advancements, a critical challenge persists: hallucination, where generated responses exhibit inconsistencies with input ~\cite{huang2024opera,leng2023mitigating,jiang2024hallucination,chen2024halc,hsun2023aligning,hwang2023llm,hchen2023mitigating,hpetryk2024aloha}.  
This phenomenon underscores fundamental challenges in multimodal understanding that demand further investigation. 
\subsection{Mitigating Hallucinations in LVLMs}  
Given the substantial time costs associated with data preparation and model training, current approaches~\cite{leng2023mitigating,huang2024opera,favero2024multi,chen2024halc,park2024convis,zou2024look,huo2024self,an2024agla,qu2024look} primarily address hallucination during inference. 
Most of them employ contrastive decoding~\cite{li2022contrastive} to rectify the original logit distribution, thereby enhancing model attention to input uni-modal information. 
Specifically, VCD~\cite{leng2023mitigating} enhances the probabilities of input-relevant tokens by contrasting logit distributions between hallucination-prone masked images and original images. 
OPERA~\cite{huang2024opera} found that partial overtrust tendencies lead to hallucinations. An over-trust logit penalty was introduced in the decoding phase to increase the focus of LVLMs on image tokens, thereby alleviating the hallucinations. 
Unlike these methods that emphasize reinforcing the focus on uni-modal information, our work focuses on mitigating hallucination by enhancing the role of multimodal interactions in decision-making of LVLMs. 
Through comparison with existing state-of-the-art methods across multiple benchmarks, we have demonstrated the effectiveness of our method. 
\subsection{Interactions of DNNs}
In recent years, an increasing number of studies have focused on quantifying the interactions among input units to analyze the representations of deep neural networks (DNNs) \cite{grabisch1999axiomatic, harsanyi1982simplified, shapley1953value,sundararajan2020shapley}.
Several studies \cite{wang2020unified,wang2021interpreting,ren2021unified} focused on interpret the adversarial transferability and  adversarial attacks of DNNs with interactions metrics. 
Other research explained the generalization power of DNNs from the perspective of interactions~\cite{Zhou2023ExplainingGP,zhang2021interpreting}.
Furthermore, certain works
\cite{zhang2021interpreting,zhang2020game} extended interactions metrics to account for multi-order and multivariate interactions, which have also been applied to explain various phenomena in DNNs \cite{zhang2021building,deng2021discovering, yao2023towards, dong2022explaining, chen2023harsanyinet}.
Different from above studies, our work focuses on analyzing the hallucination issues in Large Vision-Language Models (LVLMs) from the perspective of multimodal interactions, which still remained largely underexplored in the past literature.

\section{Multimodal Interactions in LVLMs}
\label{sec:preliminary}
In this section, we expect to investigate the roles of multimodal interactions in shaping LVLMs’ responses, with the goal of identifying potential causes of their hallucination issues. 
Specifically, we propose to first verify the existence of interactions within LVLMs' responses, then locate the scope of such interactions in LVLMs' responses, and finally evaluate the impact of such interactions on the generated responses. 
To conduct the above analyses, we designed several metrics based on the Harsanyi dividend~\cite{harsanyi1982simplified}, for which we present a brief introduction for better understanding. 

\subsection{The Harsanyi Dividend}
The Harsanyi dividend is a metric from game theory used to quantify the contribution of a coalition composed of multiple players to a game. 
Specifically, given a set of players $\mathcal{N}$ participating in a game $L$, these players may obtain a certain reward $L(\mathcal{N})$, where $L(\cdot)$ can be considered as a reward function mapping any subset of players $A \subseteq \mathcal{N}$ to a numerical value.
Under this context, each player usually does not participate in the game individually, but forms different coalitions with extra interaction effects, contributing to the final reward.
Mathematically, such effects of interactions can be uniquely measured by the Harsanyi dividend, which is defined as follows,
\begin{equation}
\label{eq2}
\mathrm{I}\left(A|\mathcal{N}\right)=\sum_{A'\subseteq A}\left(-1\right)^{|A'|-|A|}\mathrm{L}(A'),
\end{equation}

Moreover, the Harsanyi dividend is associated with the Shapley value~\cite{shapley1953value}, which adheres to several key axioms: \textit{linearity}, \textit{dummy}, \textit{symmetry} and \textit{efficiency} axioms~\cite{shapley1953value}. 
This connection provides robust theoretical support for the Harsanyi dividend, enhancing the reliability and trustworthiness of analyses built upon it.

\subsection{Quantifying the Multimodal Interactions with the Harsanyi Dividend for LVLMs}
Based on the definition of the Harsanyi Dividend, we then apply it to quantify multimodal interaction effects on the decoding process in LVLMs. 

To begin with, let us clarify the notations used for the decoding process of LVLMs.
In LVLMs, an input sample typically includes a prompt $\textbf{p}$ and an image $\textbf{v}$. 
For the token $\textbf{y}_t$ at the $t$-th step in the generated response $\textbf{y}$, LVLMs utilize information from both $\textbf{p}$ and $\textbf{v}$, as well as the previously generated tokens $\textbf{y}_{<t}$ to produce the next token. 
Formally, such a process can be expressed as follows.
\begin{equation}
\begin{array}{cc}
\widetilde{P}_{{t}}=
\mathrm{SoftMax}\left[\mathcal{M}_\theta\left(\textbf{v},\textbf{p},\textbf{y}_{<t}\right)\right], \\
\textbf{y}_t = \mathcal{S}\left(\widetilde{P}_{{t}}\right),
\end{array}
\end{equation}
where $\mathcal{M}_\theta\left(\textbf{v},\textbf{p},\textbf{y}_{<t}\right) \in \mathbb{R}^{N}$ represents any LVLMs providing logit values.
Here $N$ is the number of candidate tokens in the vocabulary set $\mathrm{B}$.
Then, in the decoding process of LVLMs, we convert $\mathcal{M}_\theta\left(\textbf{v},\textbf{p},\textbf{y}_{<t}\right)$ into the probability distribution $\widetilde{P}_{{t}}$ using the softmax function and ultimately select the token $\textbf{y}_t \in \mathrm{B}$ based on any decoding strategy $\mathcal{S}$. 

Inspired by previous research~\cite{sundararajan2020many,fryer2021shapley,wang2024diagnosing,chen2022explaining} for applying the Harsanyi dividend to DNNs, we can analogously consider the inference process of LVLMs as a game $\mathrm{L}(\cdot)$ and deem the input image $\textbf{v}$ and the input prompt $\textbf{p}$ as players.
Thus, we define the whole set of players $\mathcal{N}=\{\textbf{v},\textbf{p}\}$ and set $\mathrm{L}(\mathcal{N})^{\textbf{y}_t}= \mathcal{M}_\theta\left(\textbf{v},\textbf{p},\textbf{y}_{<t}\right)^{\textbf{y}_t}$ representing the logit value for the token $\textbf{y}_t$. 
In this way, the causal effects of the pattern $A\subseteq \{\textbf{v},\textbf{p}\}$ on the output token $\textbf{y}_t$ can be measured by the Harsanyi dividend as follows,
\begin{equation}
\label{eq2}
\mathrm{I}\left(A\right|\{\textbf{v},\textbf{p}\})^{\textbf{y}_t}=\sum_{A'\subseteq A}\left(-1\right)^{|A'|-|A|}\mathrm{L}(A')^{\textbf{y}_t},
\end{equation}
Notably, when $A=\{\textbf{v},\textbf{p}\}$, $\mathrm{I}\left(A\right|\{\textbf{v},\textbf{p}\})^{\textbf{y}_t}$ then represents the contribution of multimodal (\emph{i.e.}, image-text) interactions to the sampled token $\textbf{y}_t$.
$A'$ is a subset of $A$, which is then any one of the elements in the set $\{\{\textbf{v},\textbf{p}\},\{\textbf{v}\},\{\textbf{p}\},\emptyset\}$. 
In parallel, when $A=\{\textbf{v}\}$ or $A=\{\textbf{p}\}$, $\mathrm{I}\left(A\right|\{\textbf{v},\textbf{p}\})^{\textbf{y}_t}$ represents the contribution of each uni-modal information to the sampled token $\textbf{y}_t$. 

Based on Equation \ref{eq2}, we then design metrics to thoroughly investigate the roles of multimodal interactions on the LVLM response generation process.
For simplicity, we denote $\mathrm{I}\left(A\right|\{\textbf{v},\textbf{p}\})^{\textbf{y}_t}$ as $\mathrm{I}\left(A\right)^{\textbf{y}_t}$ in the following sections.

\subsection{Verifying the Existence of Interactions }
\label{sec3.3}
\begin{table}[!t]
\tablestyle{5pt}{1.25}
\begin{tabular}{l c c c c c}
\hline
{Datasets}
& 
\makecell{InstructBLIP\\~\cite{instructblip}}
&\makecell{LLaVA-v1.5\\~\cite{liu2023improvedllava}} & \makecell{Qwen-VL\\~\cite{bai2023qwen}} & \makecell{mPLUG-owl2\\~\cite{ye2024mplug}} \\
\hline
\hline
MME  & 0.80 & 0.59 & 2.10 & 0.07 \\
CHAIR & 0.60 & 0.06 & 0.56 & 0.16 \\
\hline
\end{tabular}
\caption{\textbf{Verifying the existence of interactions through analyzing absolute values of interaction contributions.} 
Each value represents the average absolute value 
to all generated tokens. Results reveal that all values are greater than zero, suggesting that interactions participate in the decision-making processes of LVLMs. } 
\label{tab:hopy1}
\end{table}
\noindent\fbox{
  \parbox{0.45\textwidth}{\textbf{Insight 1}: LVLMs implicitly capture multimodal interactions from input samples and leverage such interactions for decision-making to some extent.  }
}
~\\
Firstly, we propose to verify whether LVLMs already encode multimodal interactions. 
To this end, we posit that LVLMs implicitly utilize such interactions during the decision-making process.

To validate this, we analyze the absolute values of interaction contributions, $|\mathrm{I}(A)^{\textbf{y}_t}|$, with $A=\{\textbf{v},\textbf{p}\}$. If this value is greater than zero, it indicates that multimodal interactions influence the generation of LVLM responses.

As shown in Table~\ref{tab:hopy1}, we calculated the mean absolute value of $|\mathrm{I}(A)^{\textbf{y}_t}|$ across all sampled tokens. 
Results demonstrate consistent positive values across multiple benchmarks including MME~\cite{fu2023mme} and CHAIR~\cite{rohrbach2018object}. 
This empirical evidence confirms that LVLMs do implicitly capture image-text interactions during response generation to some extent.

\subsection{Verifying the Scope of Interactions}
\label{sec3.4}
\noindent\fbox{
  \parbox{0.45\textwidth}{\textbf{Insight 2}: LVLMs exhibit a tendency to primarily apply their understanding of multimodal interactions to the generation of a few key tokens, rather than uniformly across all response tokens. 
}
}
~\\
Based on the results in Section \ref{sec3.3}, we further propose to locate the
scope of such interactions in LVLMs’ responses.
In other words, we expect to analyze how LVLMs leverage image-text interactions to formulate responses during the answer-generation phase. 
Specifically, LVLM-generated answers are typically composed of keywords and contextual connectives. 
Keywords indicate question-relevant content, while contextual connectives integrate these words and convert them into language expressions.
We hypothesize that image-text interactions primarily influence the reasoning process behind keywords generations, shaping the core content of responses. 
\begin{figure}[!t]
    \centering
    \includegraphics[width=1.0\linewidth, trim = 0 0 0 7,clip]{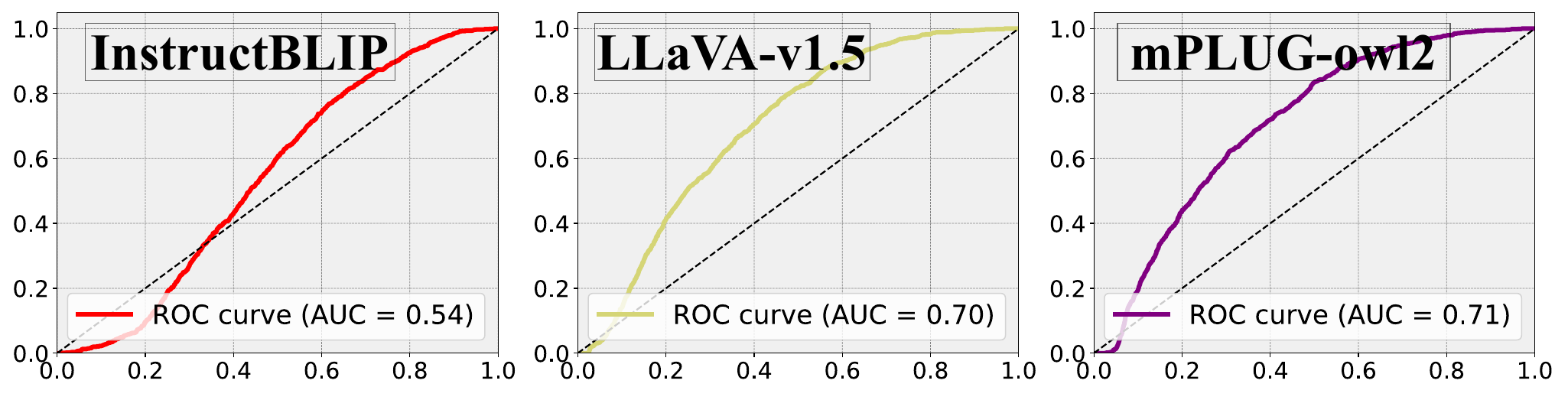}
    \caption{\textbf{Verifying the scope of interactions using the Receiver Operating Characteristic (ROC) curve.}  
Results (AUC $>$ 0.5) show that there exists a moderate class separation between keywords tokens and contextual connectives tokens based on the variance of multimodal interactions. These findings suggest that multimodal interactions primarily impact the the generations of keywords in LVLMs' responses. }
    \label{fig:hopy2}
\end{figure}
\begin{figure*}[!ht]
    \centering
\includegraphics[width=1.0\linewidth,trim = 0 0 0 10,clip]{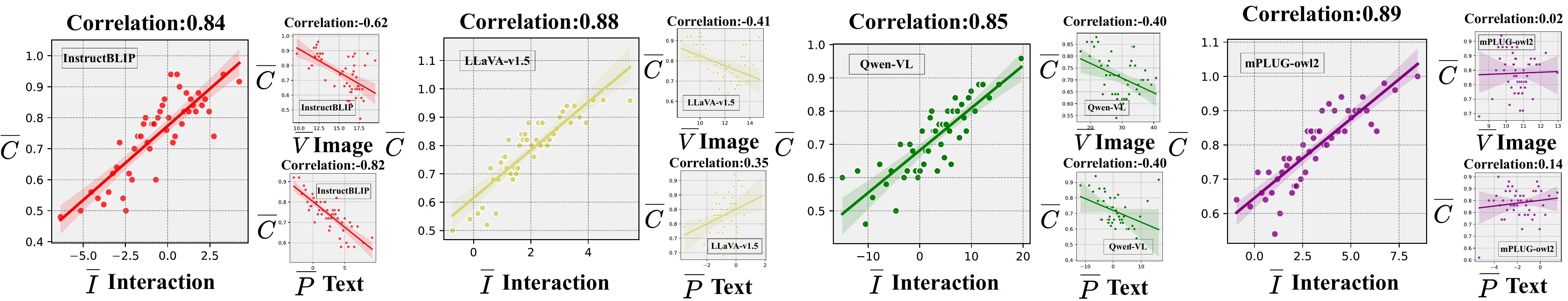}
    \caption{
\textbf{Verifying the impact of interactions through the Pearson correlation coefficient. }
The set of figures for each LVLM illustrates the correlation between the accuracy and the multimodal interaction contributions, the correlation between the accuracy and the visual modality contributions, and the correlation between the accuracy and the textual modality contributions on MME~\cite{fu2023mme} benchmark. 
    Each point corresponds to statistics within a bin $\mathbf{b}$ of 50 tokens, including the proportion of correct tokens $\overline{C}_{\mathbf{b}}$, the average contribution of multimodal interactions $\overline{I}_{\mathbf{b}}$, the average contribution of visual modality $\overline{V}_{\mathbf{b}}$ and the average contribution of textual modality $\overline{P}_{\mathbf{b}}$. 
    Results indicate that as the contribution of multimodal interactions increases, the proportion of correct tokens rises, resulting in higher accuracy. 
    Additionally, the influence of multimodal interactions in LVLMs shows a stronger positive correlation with prediction accuracy than that of uni-modal information, indicating that the contribution of multimodal interactions plays a more critical role in ensuring accurate responses. 
}
\label{fig:h1-1}
\end{figure*}

The key challenge here is to quantitatively measure the involvement of multimodal interactions in the generation of each token. 
To this end, we propose to utilize the variance of multimodal interaction effects
$\mathcal{D}_{\textbf{y}_t}(\mathrm{I}(A)^{\textbf{y}_t})$ as a measurement for the level of interaction engagement at the $t$-th step. 
Specifically, when the variance $\mathcal{D}_{\textbf{y}_t}(\mathrm{I}(A)^{\textbf{y}_t})$ presents a small value, it indicates that multimodal interactions had almost the same influence on the candidate tokens in the vocabulary for the $t$-th step. 
In other words, multiple candidate tokens exhibit similar $\mathrm{I}(A)^{\textbf{y}_t}$, and the model's token selection becomes interaction-agnostic, implying weak engagement of interactions.
By contrast, if $\mathcal{D}_{\textbf{y}_t}(\mathrm{I}(A)^{\textbf{y}_t})$ shows a large value, it indicates that multimodal interactions provides meaningful guidance to a few candidate tokens among the whole vocabulary set, implying strong engagement in the token selection. 
In this way, we hypothesize that the generations of keywords should present higher values of $\mathcal{D}_{\textbf{y}_t}(\mathrm{I}(A)^{\textbf{y}_t})$, compared to contextual connectives.

To validate this, we sample a subset in CHAIR \cite{rohrbach2018object} following OPERA~\cite{huang2024opera} and conduct experiments to analyze the differences in $\mathcal{D}_{\textbf{y}_t}(\mathrm{I}(A)^{\textbf{y}_t})$ between keywords and contextual connectives in LVLMs' responses. 
To localize keywords, we tag the part-of-speech of generated words based on spaCy~\cite{honnibal2017spacy}.
We argue that nouns typically denote objects, while adjectives, numbers, and adverbs specify the attributes of those objects; verbs and adpositions establish inter-object relationships. 
Therefore, we regard such words as keywords and others as contextual connectives. 
In this way, we treat keywords as positive class instances and construct Receiver Operating Characteristic (ROC) curves by using variance $\mathcal{D}_{\textbf{y}_t}(\mathrm{I}(A)^{\textbf{y}_t})$ as the classification threshold. 

As shown in \cref{fig:hopy2}, the Area Under the Curve (AUC) values of various LVLMs exceed 50\%. 
Such results reveal that there exists a moderate class separation between keywords and contextual connectives based on the variance of multimodal interactions $\mathcal{D}_{\textbf{y}_t}(\mathrm{I}(A)^{\textbf{y}_t})$. 
This empirical evidence confirms that interactions in LVLMs are primarily focused on the reasoning of keywords tokens, rather than being applied uniformly to the entire autoregressive process.

\subsection{Verifying the Impact of Interactions}
\noindent\fbox{
  \parbox{0.45\textwidth}{
  \textbf{Insight 3}: The understanding of multimodal interactions in LVLMs positively influences the quality of generated responses, with stronger interactions leading to greater accuracy.  
}
}
~\\
Based on the results in Section \ref{sec3.4}, we expect to further evaluate the impact of multimodal interactions on the generation of keywords tokens.
We hypothesize that such multimodal interactions should bring a positive impact on keywords generations, with stronger interactions leading to more accurate responses.

It is challenging to quantify the relationship between the strength of multimodal interactions for keyword tokens and the overall accuracy of the response. To address this, we propose conducting experiments on the MME~\cite{fu2023mme} benchmark for convenient verification.
In MME, the keywords in the responses are limited to ``yes" or ``no", which are also used to determine the accuracy of the responses. This setup allows us to measure the strength of multimodal interactions for keyword tokens using $\mathrm{I}(A)_r^{\textbf{y}_t}$, where $\textbf{y}_t \in \{\textit{yes}, \textit{no}\}$ and $A=\{\textbf{v},\textbf{p}\}$, for each response $r$. The appearance of the \textit{yes}/\textit{no} keywords directly indicates the binary correctness of the response.
Furthermore, to better verify the trend of the positive impact of multimodal interactions, we propose to partition all the generated responses on MME into multiple bins based on the value of $\mathrm{I}(A)_r^{\textbf{y}_t}$ ($\textbf{y}_t \in \{yes, no\}$) sorted in ascending order, with each bin containing $Q$ samples.
Then, we calculate the average contribution of interactions within each bin $\mathbf{b}$, denoted by $\overline{I}_\mathbf{b}=\frac{1}{Q}\sum_{r \in \mathbf{b}}\mathrm{I}(A)_r^{\textbf{y}_t}$.
We denote $\overline{C}_{\mathbf{b}}$ as the ratio of correct responses in each bin $\mathbf{b}$.
Thus, we can analyze the correlation between $\overline{I}_\mathbf{b}$ and $\overline{C}_\mathbf{b}$ among different bins, allowing us to verify trends in the positive impact of multimodal interactions on response accuracy.

As shown in \cref{fig:h1-1}, we observe a significant monotonic relationship, where higher strength of multimodal interactions corresponds to improved response accuracy. This highlights the effectiveness of multimodal interactions in guiding keyword token selections.
To fully complement our analysis for contrast, we also quantified uni-modal contributions on each keyword token $\textbf{y}_t$ through calculating $\mathrm{I}\left(\{\textbf{v}\}\right)_r^{\textbf{y}_t}$ and $\mathrm{I}\left(\{\textbf{p}\}\right)_r^{\textbf{y}_t}$. 
Using the same binning procedure described earlier, we calculated the average contribution of visual modality and textual modality, \emph{i.e.}, $\overline{V}_\mathbf{b}=\frac{1}{Q}\sum_{r \in \mathbf{b}}\mathrm{I}(\{\mathbf{v}\})_r^{\textbf{y}_t}$ and $\overline{P}_\mathbf{b}=\frac{1}{Q}\sum_{r \in \mathbf{b}}\mathrm{I}(\{\mathbf{p}\})_r^{\textbf{y}_t}$ for each bin $\mathbf{b}$. 
In \cref{fig:h1-1}, the correlation of $\overline{I}$ and $\overline{C}$, the correlation of $\overline{V}$ and $\overline{C}$, and the correlation of $\overline{P}$ and $\overline{C}$ are compared. 
Notably, our quantitative analysis demonstrates a more positive correlation between multimodal interactions and response accuracy compared to uni-modal information alone. 
These findings suggest that image-text interactions plays a more substantial role in generating reliable responses.

\section{INTER: Interaction Guidance Sampling}
\label{sec:method}
Based on the obtained insights, we find that LVLMs exhibit a human-like, though less pronounced, understanding of multimodal interactions, which motivated us to develop efficient methods to reduce hallucinated responses in LVLMs.
To this end, we further propose a simple yet effective method named Interaction Guidance Sampling (INTER). 
The approach comprises two components: the Interactive Guided Locator and the Interaction Probability Modifier. 
Specifically, the Interactive Guided Locator is designed to localize keywords in generated responses.
Such words typically indicate visual information that is relevant to questions (\emph{e.g.}, objects, object relations, or object attributes).
During the keyword generation phase, the Interaction Probability Modifier then guides LVLMs to sample tokens that exhibit stronger reliance on the image-text interactions.
Sampling tokens based on the interactions, INTER prevents the interference of the question-irrelevant information, thus reducing the hallucination in LVLMs.

\begin{figure}
    \centering
    \includegraphics[width=1.0\linewidth]{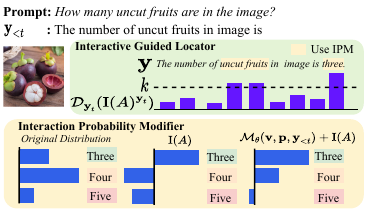}
    \caption{\textbf{Overview of INTER.}  
    The Interactive Guided Locator (IGL) uses variance of multimodal interactions to identify keywords of interest.     
    The Interaction Probability Modifier (IPM) uses multimodal interactions to guide the model to sample tokens responsive to visual information related to the question. }
    \label{fig:method}
\end{figure}

\noindent \textbf{Interactive Guided Locator. }LVLMs primarily apply the understanding of multimodal interactions to the generation of a few keywords tokens.
To this end, we propose the Interactive Guided Locator (IGL) method to indicate such tokens. 
Following \cref{sec3.4}, the module calculates the variance of interactions $\mathcal{D}_{\textbf{y}_t}(\mathrm{I}(A)^{\textbf{y}_t})$ at $t$-th step during answer synthesis.
As shown in \cref{fig:method}, if the variance exceeds the predefined threshold $k$, IGL designates $\beta=\mathbf{1}_{\{\mathcal{D}_{\textbf{y}_t}(\mathrm{I}(A)^{\textbf{y}_t}) > k\}}$, which signifies that the token $\textbf{y}_t$ is a keyword in the answer. 
Conversely, if $\mathcal{D}_{\textbf{y}_t}(\mathrm{I}(A)^{\textbf{y}_t})$ is less than $k$, we set $\beta$ to $\mathbf{0}$ showing $\textbf{y}_t$ is contextual connective.

The Interactive Guided Locator (IGL) localizes keywords in sentences and guides the Modifier only applying interactions enhancement to these tokens.
In this way, IGL prevents interference from image-text interactions to contextual connectives, thereby preserving the linguistic coherence of generated answers.

\noindent \textbf{Interaction Probability Modifier. } 
The understanding of multimodal interactions in LVLMs positively influences the quality of generated responses.
Based on the insight, we present the Interaction Probability Modifier (IPM) to guide LVLMs sampling tokens. 
In this way, tokens that are more influenced by image-text interactions are sampled by LVLMs, thereby improving the relevance of answers to multimodal inputs. 

Specifically, we utilize Harsanyi dividend~\cite{harsanyi1982simplified} to explicitly decouple interactions and treat interactions as prior knowledge to adjust the original logit distribution, as shown in \cref{fig:method}. 
For each candidate token $\textbf{y}_t$, we can obtain the multimodal interaction contribution $\mathrm{I}(A)^{\textbf{y}_t}$ with $A=\{\textbf{v},\textbf{p}\}$, as well as the original logit $\mathcal{M}_\theta\left(\textbf{v},\textbf{p},\textbf{y}_{<t}\right)^{\textbf{y}_t}$. 
Then, we use $\mathrm{I}(A) \in R^{N}$ to modify the logit of each candidate token and generate sampling probabilities, formulated as follows: 
\begin{equation}
  \widetilde{P}_{{t}} = \mathrm{SoftMax}\left[\mathcal{M}_\theta\left(\textbf{v},\textbf{p},\textbf{y}_{<t}\right)+\text{I}\left(A\right)\right].  
\end{equation}
By doing so, IPM enhances the dependence of LVLMs on multimodal interactions. 
Therefore, the generation of information unrelated to the input is suppressed, reducing hallucinations in LVLMs. 

\textbf{Overall Mechanism. }
The INTER employs multimodal interaction guidance sampling for keywords, formulated as:
\begin{equation}
\widetilde{P}_{{t}} = \mathrm{SoftMax}\left[\mathcal{M}_\theta\left(\textbf{v},\textbf{p},\textbf{y}_{<t}\right) + \beta\cdot\mathrm{I}(A)\right],
\end{equation}
by using a threshold to control $\beta$ for locating key steps, we help the model focus more on image-text interactions during keywords generations. 

\section{Experiment}

\begin{figure*}[!t]
    \centering
\includegraphics[width=1.0\linewidth,trim = 0 0 0 30,clip]{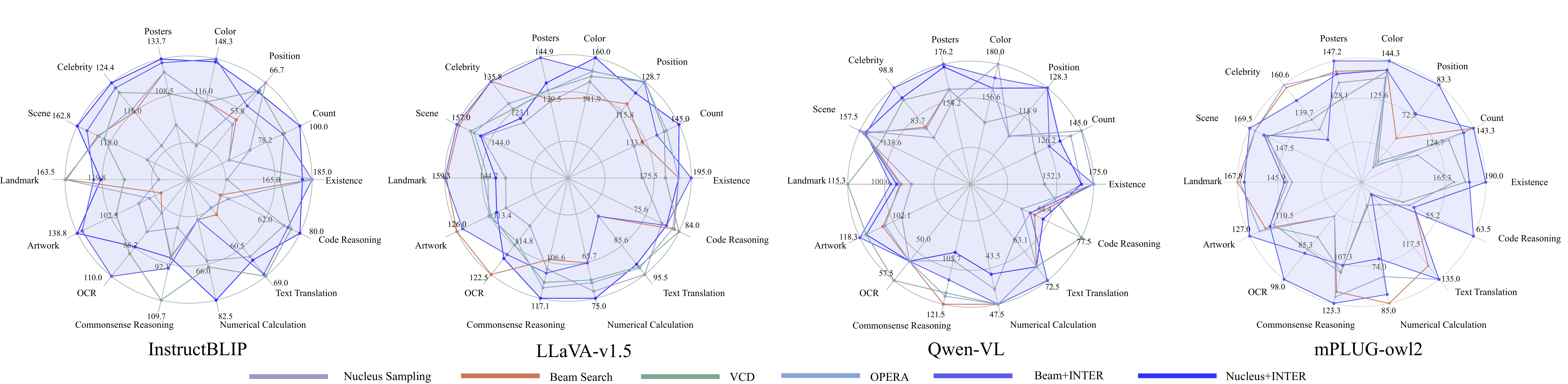}
    \caption{\textbf{The scores of 14 subtasks on MME benchmark~\cite{fu2023mme}.} The radar charts show that the areas filled in purple are significantly larger, suggesting that using INTER results in noticeable improvements. } 
    \label{fig:mme_fig}
\end{figure*}
\begin{table*}[!t]
\tablestyle{5pt}{1.2}
\begin{tabular}{lcccccccc}
\multirow{2}{*}{method}&   
\multicolumn{2}{c}{InstructBLIP~\cite{instructblip}} & 
\multicolumn{2}{c}{LLaVA-v1.5 (7B)~\cite{liu2023improvedllava}}& 
\multicolumn{2}{c}{Qwen-VL~\cite{bai2023qwen}} & 
\multicolumn{2}{c}{mPLUG-owl2~\cite{ye2024mplug}}  \\
& COCO~\cite{lin2014microsoft} & AOKVQA~\cite{schwenk2022okvqa} & COCO~\cite{lin2014microsoft} & AOKVQA~\cite{schwenk2022okvqa}& COCO~\cite{lin2014microsoft} & AOKVQA~\cite{schwenk2022okvqa}  & COCO~\cite{lin2014microsoft} & AOKVQA~\cite{schwenk2022okvqa}\\ \hline
\vb Nucleus~\cite{holtzman2019curious} 
&  80.1 
& 78.5 
& 79.7 
& 79.1 
& 81.7
&83.2
&  80.4 
& 78.0\\
\gr
\cb Nucleus+INTER  & 
83.3 ($\uparrow$3.2) & 
82.4 ($\uparrow$3.9) & 
85.7 ($\uparrow$6.0) & 
82.6 ($\uparrow$3.5) & 
86.2 ($\uparrow$4.5) &
86.1 ($\uparrow$2.9) &
81.9 ($\uparrow$1.5) & 
79.1 ($\uparrow$1.1)
\\
 
 \vb Beam~\cite{graves2012sequence, sutskever2014sequence,boulanger2013audio} 
 & 81.9 &  81.1
 & 84.9 & 84.3 
 & 83.4& 85.0
 &83.3 & 82.3  \\
\gr
 \cb Beam+INTER 
 & 84.6 ($\uparrow$2.7) &  83.6 ($\uparrow$2.5) 
 & 85.5 ($\uparrow$0.6) & 84.9 ($\uparrow$0.6) 
 & 86.1 ($\uparrow$2.7) &86.4 ($\uparrow$1.4) 
 &83.7 ($\uparrow$0.4)& 82.3 ($\uparrow$0.0)  \\
 \vb VCD\textsuperscript{*}~\cite{leng2023mitigating} 
 &81.4 & 81.0 
 & 84.5 & 82.3 
 &  86.0& 86.4
 & 82.3 & 79.2\\
\gr
\cb VCD\textsuperscript{*}+INTER & 
82.9 ($\uparrow$1.5) & 
80.8 ($\downarrow$0.2) & 
85.6 ($\uparrow$1.1) & 
83.0 ($\uparrow$0.7) & 
86.3 ($\uparrow$0.3) &
86.4 ($\uparrow$0.0) &
82.8 ($\uparrow$0.5) & 
79.3 ($\uparrow$0.1) \\
\vb OPERA\textsuperscript{\dag}~\cite{huang2024opera} 
&84.7 &83.7  
& 85.3 & 84.1 
& 83.4& 85.1
& 83.4 & 82.1 \\
\gr
\cb OPERA\textsuperscript{\dag}+INTER & 
85.2 ($\uparrow$0.5) & 
85.0 ($\uparrow$1.3) & 
85.8 ($\uparrow$0.5) &
84.8 ($\uparrow$0.7) & 
83.5 ($\uparrow$0.1)&
84.4 ($\downarrow$0.7)&
83.4 ($\uparrow$0.0)&
82.3 ($\uparrow$0.2)\\
\end{tabular}
\caption{\textbf{The average F1-score on the POPE benchmark~\cite{li2023pope}.} $\uparrow$ means that higher values indicate lower hallucination levels. The results indicate that applying INTER calibration, the models showed a reduction in hallucinations. }
\label{tab:pope1}
\end{table*}

\label{sec:exp}
In this section, we first describe our experimental settings. 
Subsequently, we present the model’s performance before and after the application of INTER across various decoding strategies to demonstrate the effectiveness of INTER. 
Following this, we conduct a parameter analysis and evaluate the performance of INTER across LVLMs with varying parameter scales. 
Additional results into the performance of INTER can be found in the supplementary materials. 
\subsection{Experimental Settings}
\textbf{Models. }We conduct experiments on multiple representative LVLMs to demonstrate the generalization ability of INTER. 
Specifically, our experiments includes the 7B version of Qwen-VL~\cite{bai2023qwen}, InstructBLIP~\cite{instructblip} and mPLUG-owl2~\cite{ye2024mplug}, the 7B and 13B versions of LLaVA-v1.5~\cite{liu2023improvedllava}, and 1-26B versions of InternVL2~\cite{chen2024far,chen2024internvl}. 

\textbf{Benchmarks. }To rigorously validate INTER's enhanced cross-modal comprehension capabilities, we conducted extensive evaluations across Visual Question Answering (VQA) and image captioning tasks. 
For VQA assessment, we employed four authoritative benchmarks: Polling-based Object Probing Evaluation (POPE)~\cite{li2023evaluating} for object hallucination analysis, along with three benchmarks for comprehensive evaluations: MME~\cite{fu2023mme}, MM-Bench~\cite{liu2023mmbench} and MMStar~\cite{chen2024we}. 
The captioning performance was assessed through Caption Hallucination Assessment with Image Relevance (CHAIR)~\cite{rohrbach2018object} to quantify object-level hallucination. 
Additionally, LLaVA-Bench~\cite{llava_bench_in_the_wild} was utilized to analyze hallucinations in open-ended, real-world scenarios. 
Comprehensive details regarding these benchmarks are provided in the supplementary materials. 

\textbf{Metrics. }In the POPE~\cite{li2023pope} benchmark, we implement three question sampling strategies for each dataset, reporting the average F1 score as the primary evaluation metric. 
Under the MME~\cite{fu2023mme} evaluation, we compute both the total score and perception score, alongside the performance across various subtasks, according to VCD~\cite{leng2023mitigating}. 
For the MM-Bench~\cite{liu2023mmbench} and MMStar~\cite{chen2024we} benchmarks, we focus on overall performance metrics, while comprehensive analyses of the subtasks are provided in the supplementary materials. 
Based on OPERA~\cite{huang2024opera}, we reported two metrics: $CHAIR_S (C_S)$ and $CHAIR_I(C_I)$ on CHAIR~\cite{huang2024opera}, which assess the degree of hallucinations at both the sentence and image levels. 
For the LLaVA-Bench~\cite{llava_bench_in_the_wild} open-ended generation tasks, we use GPT-4o~\cite{hurst2024gpt} to score a 1-10 scale from two dimensions: semantic accuracy and detail richness.  

\textbf{Baselines. }
We conduct evaluations across five decoding strategies to validate the effectiveness of INTER on hallucination mitigation: Nucleus Sampling~\cite{holtzman2019curious} ($p=1.0$), Beam Search~\cite{graves2012sequence, sutskever2014sequence,boulanger2013audio} ($N_{beam}=5$), Greedy Search~\cite{song2024good} and two state-of-the-art methods VCD\textsuperscript{*}~\cite{leng2023mitigating} and OPERA\textsuperscript{\dag}~\cite{huang2024opera}. 
\textsuperscript{*} and \textsuperscript{\dag} represent correction based on Nucleus Sampling and Beam Search. 
INTER-enhanced strategies are denoted with `+INTER', which substitute visual inputs of $\{\textbf{p}\}$ with random noise while initialize text inputs of $\{\textbf{v}\}$ with empty text inputs.  
All experiments maintain hyperparameters from VCD~\cite{leng2023mitigating} and OPERA~\cite{huang2024opera} implementations for fair comparisons. 
Quantitative results represent averages over five runs. 
\subsection{Experimental Results}
 \begin{table*}[t!]

\tablestyle{3pt}{1.3}
\begin{tabular}{lccccccccc}
model& benchmark&Nucleus& Nucleus+INTER&Beam& Beam+INTER&VCD\textsuperscript{*}& VCD\textsuperscript{*}+INTER&OPERA\textsuperscript{\dag} &OPERA\textsuperscript{\dag}+INTER\\\hline
\multirow{2}{*}{\makecell{LLaVA-v1.5 (7B)\\~\cite{liu2023improvedllava}}}&  MM-Bench~\cite{liu2023mmbench} $\uparrow$&57.3&\textbf{62.6}	&65.1	&\textbf{65.1}&62.5&\textbf{62.9}&65.0&\textbf{65.0	}\\
    &MMStar~\cite{chen2024we} $\uparrow$ &29.3&\textbf{31.9}	&31.1	&\textbf{31.7}&30.3&\textbf{31.1}&31.4&\textbf{32.9}	\\\hline
    \multirow{2}{*}{\makecell{mPLUG-owl2\\~\cite{ye2024mplug}}}&  MM-Bench~\cite{liu2023mmbench} $\uparrow$&57.0&\textbf{61.4}	&63.5	&\textbf{63.7}&59.2&\textbf{59.5}&63.4&\textbf{63.6}	\\
    &MMStar~\cite{chen2024we} $\uparrow$&30.5&\textbf{32.3}&30.7	&\textbf{30.5}&31.5&\textbf{32.3}&30.5&\textbf{30.9}	\\
\end{tabular}
\caption{\textbf{Validation of INTER on MM-Bench~\cite{liu2023mmbench} and MMStar~\cite{chen2024we}.} $\uparrow$ means that higher values indicate lower hallucination levels.}
\label{tab:mmbench+MMStar}
\end{table*}

\textbf{Result on POPE~\cite{li2023pope}. }We conducted comparative experiments on POPE to demonstrate the effectiveness of INTER in enhancing LVLMs' performance on object existence tasks. 
As demonstrated in Table~\ref{tab:pope1}, INTER achieves consistent improvements across all datasets in POPE, with maximum mean F1-score enhancement reaching 7.5\%. Notably, INTER-enhanced Nucleus Sampling outperforms VCD, while `Beam+INTER' surpasses OPERA's performance. 
These comparative results validate that emphasizing cross-modal interaction yields superior performance over uni-modal enhancement methods.

\textbf{Result on MME~\cite{fu2023mme}. }Quantitative analysis through 14 subtasks of MME~\cite{fu2023mme} reveals the task-specific advantages of INTER. 
As visualized in \cref{fig:mme_fig}, our method outperforms baseline methods in most subtasks, particularly excelling in subtasks related to scene text recognition and fine-grained attribute identification. 
Moreover, INTER demonstrates a 343.7-point absolute improvement in the total score of all 14 subtasks compared to Nucleus Sampling, with more detailed results provided in the supplementary material. 


\textbf{Result on MM-Bench~\cite{liu2023mmbench} and MMStar~\cite{chen2024we}. }MM-Bench and MMStar evaluate LVLMs 
through various subtasks including compositional reasoning and fine-grained perception. 
As shown in \cref{tab:mmbench+MMStar}, INTER achieves consistent accuracy improvements across these benchmarks.

\begin{table}[!t]
\tablestyle{2pt}{1.3}
\begin{tabular}{llcccccc}
\multirow{2}{*}{\makecell{Max \\Token}}& \multirow{2}{*}{method} & \multicolumn{2}{c}{\makecell{InstructBLIP\\~\cite{instructblip}}} & \multicolumn{2}{c}{\makecell{LLaVA-v1.5 \\(7B)~\cite{liu2023improvedllava}}} &\multicolumn{2}{c}{\makecell{mPLUG-owl2\\~\cite{ye2024mplug}}}\\
&&$C_S\downarrow$&$C_I\downarrow$&$C_S\downarrow$&$C_I\downarrow$&$C_S\downarrow$&$C_I\downarrow$ \\\hline
&\vb Nucleus~\cite{holtzman2019curious}& 29.0	&15.3	&24.4	&9.4&	26.2&	10.9\\
\gr&\cb Nucleus+INTER  &\textbf{26.2}	&\textbf{10.0}	&\textbf{19.8}	&\textbf{6.7}&	\textbf{24.8}	&\textbf{9.4}\\
& \vb Beam~\cite{graves2012sequence, sutskever2014sequence,boulanger2013audio} & 21.4&	7.2	&19.4&	6.2&	21.6&	7.6\\
\gr&\cb Beam+INTER &\textbf{21.0}	&\textbf{6.4}&	\textbf{17.8}&	\textbf{5.9}	&\textbf{21.2}&	\textbf{7.6}\\

64&\vb VCD\textsuperscript{*}~\cite{leng2023mitigating} &33.0&	11.8&	24.4&	8.0	&24.0&	9.5 \\
\gr&\cb VCD\textsuperscript{*}+INTER &	\textbf{31.2}&	\textbf{11.4}	&\textbf{21.0}&\textbf{7.2}	&\textbf{21.2}&	\textbf{8.3}\\
&\vb OPERA\textsuperscript{\dag}~\cite{huang2024opera}  &19.9&	6.8&	19.0&	6.6	&21.0	&7.8  \\
\gr&\cb OPERA\textsuperscript{\dag}+INTER & \textbf{18.4}&	7.9	&\textbf{19.0}&	\textbf{6.3}&	\textbf{20.0}	&\textbf{7.4} \\ \hline
&\vb Nucleus~\cite{holtzman2019curious}& 61.0	&28.4	&54.0	&16.1&	60.8	&20.1\\
\gr&\cb Nucleus+INTER  &\textbf{59.0}	&\textbf{20.8}	&\textbf{51.8}	&\textbf{14.1}&	\textbf{59.4}&	\textbf{19.3}\\
 &\vb Beam~\cite{graves2012sequence, sutskever2014sequence,boulanger2013audio} & 55.6	&15.8	&48.8&	13.9	&56.4	&17.9\\
\gr&\cb Beam+INTER &\textbf{55.4}	&\textbf{13.1}	&\textbf{46.4}	&\textbf{13.4}	&\textbf{53.4}&	\textbf{17.2}\\
512&\vb VCD\textsuperscript{*}\cite{leng2023mitigating} &58.6	&19.2	&53.8	&16.0&	62.8&	20.5 \\
\gr&
\cb VCD\textsuperscript{*}+INTER &	\textbf{56.2}	&\textbf{18.8}	&56.0&	\textbf{15.7}	&\textbf{60.4}	&\textbf{20.5} \\
&\vb OPERA\textsuperscript{\dag}~\cite{huang2024opera}  &48.7&	13.5	&45.4	&13.8&	55.2	&16.1  \\
\gr&
\cb OPERA\textsuperscript{\dag}+INTER & \textbf{42.2}	&18.8&	47.0&	\textbf{13.6}	&\textbf{52.5}	&\textbf{15.9}\\\hline
\end{tabular}
\caption{\textbf{Result on CHAIR~\cite{rohrbach2018object}.} $\downarrow$ means that lower values indicate lower hallucination levels. }
\label{tab:chair64}
\end{table}
\textbf{Result on CHAIR~\cite{rohrbach2018object}.  }To evaluate INTER's hallucination mitigation capability in image captioning, we conduct experiments on 500 randomly sampled instances from the CHAIR benchmark with caption lengths 64 and 512 tokens in \cref{tab:chair64}. Both $C_S$ and $C_I$ metrics show lower values indicating reduced hallucinations, where INTER achieves t he highest reduction of 34.6\% and 18.9\% on $C_S$ and $C_I$ respectively compared to baseline methods. 
\subsection{Further Analysis}

\begin{figure}
    \centering
    \includegraphics[width=1.0\linewidth]{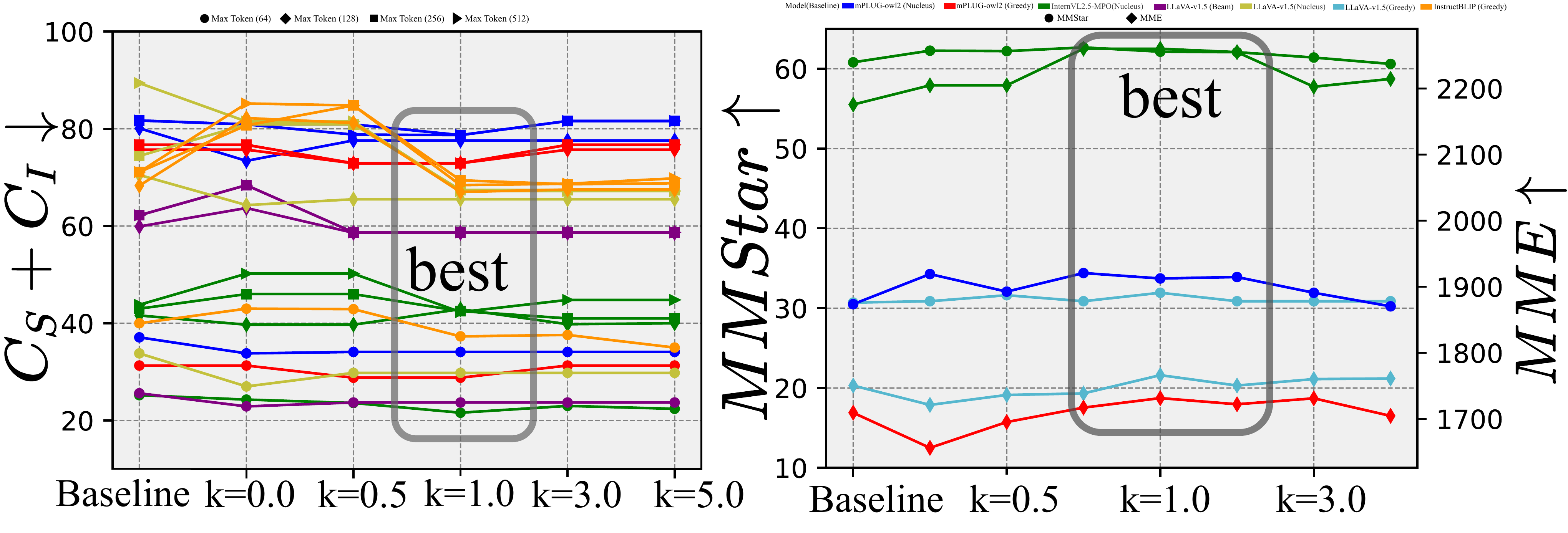}
    \caption{The robustness of the selection of hyperparameter (k). }
    \label{fig:5-3-1}
\end{figure}

\textbf{The Robustness of the Selection of Hyperparameter (k). }As shown in \cref{fig:5-3-1}, when $k=1$, various LVLMs achieve relatively good performance on different tasks. Furthermore, within the range of $k\in[0.5, 1.5]$, the performance of the LVLM remains stable. 
Such results demonstrate that INTER exhibits robustness to the selection of $k$.

\textbf{Effectiveness at Different Parameter Scales. }Our evaluation establishes INTER's cross-scale adaptability through validation across scales spanning 1B to 26B parameters. 
As shown in Table~\ref{tab:db}, the MME~\cite{fu2023mme} benchmark reveals consistent performance improvements across diverse scales, demonstrating scaling-agnostic generalization capabilities. 


    \begin{table}[!t]
\tablestyle{2.5pt}{1.3}
\begin{tabular}{lcccccc}
 method& \multicolumn{4}{c}{InternVL2~\cite{chen2024far,chen2024internvl}}&\multicolumn{2}{c}{LLaVA-v1.5~\cite{liu2023improvedllava}} \\
 &1B&4B&8B&26B&7B&13B\\\hline
\vb Nucleus~\cite{holtzman2019curious} &1689.0&1914.7&2040.9&	1950.3&1502.2&1637.2\\
\gr\cb Nucleus+INTER&\textbf{1721.6}&\textbf{1985.1}&\textbf{2131.5}&\textbf{2047.0}&\textbf{1731.6}&\textbf{1690.3}\\
\vb Beam~\cite{graves2012sequence, sutskever2014sequence,boulanger2013audio}  &1715.8&2074.3	& 2145.8&2219.0&1707.2&1760.5	\\
\gr\cb Beam+INTER &	\textbf{1743.1}&\textbf{2075.7}	& \textbf{2179.7}	&\textbf{2240.7}&\textbf{1744.0}&\textbf{1768.9}\\
\end{tabular}
\caption{\textbf{Effectiveness of INTER on different parameter scales.} }
\label{tab:db}
\end{table}

\section{Conclusion}
In this paper, we confirmed the unexplored phenomenon of the existence, scope, and effects of multimodal interactions in the entire decision-making process of LVLMs. 
Inspired by this, we propose Interaction Guidance Sampling, a training-free approach that first mitigates hallucinations from the perspective of enhancing the reliance of LVLMs on multimodal interactions. 
Extensive experiments demonstrate that by reducing the sampling of input-irrelevant information, INTER effectively mitigated hallucinations in the responses of LVLMs. 
In summary, this research provides a new perspective for mitigating hallucinations in LVLMs, shedding new light on the development of the field. 

\vspace{3mm}
\noindent\textbf{Acknowledgements.} This work was supported by the National Natural Science Foundation of China under Grant No. U22A2098; the Key Science and Technology Development Plan of Jilin Province under Grant
No. 20240302078GX. This work was supported by Alibaba Group through Alibaba Research Intern Program.

{
    \small
    \bibliographystyle{ieeenat_fullname}
    \bibliography{main}
}
\clearpage

\section{Details of the Benchmarks}

\textbf{POPE. }The Polling-based Object Probing Evaluation (POPE)~\cite{li2023evaluating} utilizes images sampled from several datasets, including MSCOCO~\cite{lin2014microsoft}, A-OKVQA~\cite{schwenk2022okvqa}, and GQA~\cite{hudson2019gqa}. 
Every question in POPE is \textit{"Is there a $<$object$>$ in the image?"}. 
For each dataset, it incorporates random, popular, and adversarial question sampling strategies to sample $<$object$>$ and create three partitions. 
Random represents randomly selecting an object from the candidate object set.
Popular means selecting the objects that occur more frequently. 
Adversarial refers to select objects that have a high co-occurrence frequency with the objects in the image. 
Therefore, the adversarial partition is the most challenging, as hallucinations are often caused by a high co-occurrence frequency between objects. 

\textbf{MME. }MME~\cite{fu2023mme} evaluates LVLMs using 14 subtasks from the perspectives of perception and cognition. 
There are four subtasks for the evaluation of the cognition ability, including commonsense reasoning, numerical calculation, text translation, and code reasoning. 
The remaining subtasks are used to evaluate perceptual abilities from the perspectives of coarse grained recognition, fine grained recognition, and OCR. 
Each image corresponds to two questions with opposing answers. 
For each subtask, the score of LVLMs is represented by the proportion of all questions answered correctly, as well as the proportion of both questions for each image answered correctly. 

\textbf{MM-Bench. }MM-Bench~\cite{liu2023mmbench} 
employs 20 subtasks to evaluate LVLMs in detail. 
These 20 subtasks are further divided into six perspectives: `Coarse Perception (CP)', `Cross-instance Fine-grained Perception (FP-C)', `Single-instance Fine-grained Perception (FP-S)', `Attribute Reasoning (AR)', `Logic Reasoning (LR)', and `Relation Reasoning (RR)'. 
For each sample, MM-Bench sets several options and requires the LVLMs to return one of them. 
The template for each question is \textit{`Answer with the option's letter from the given choices directly.'}. 
More importantly, MM-Bench creates questions with the same content but differing option sequences by repeatedly rotating the order of them.  
Then, for each sample, the accuracy across all orders is collected, and if all are answered correctly, the LVLMs score for that sample. 
Therefore, MM-Bench's evaluation of LVLMs is more rigorous and is not influenced by the order of the options. 

\textbf{MMStar. }Like MM-Bench, MMStar~\cite{chen2024we} also establishes multiple subtasks and categorizes them into six perspectives: `Coarse Perception (CP)', `Fine-Grained Perception (FP)', `Instance Reasoning (IR)', `Logical Reasoning (LR)', `Science \& Technology (ST)' and `Math (MA)'. 
Every aspect have three subtasks. 
But the difference is that MMStar uses a four-tier filtering mechanism to select 1,500 elite samples from an initial pool of 22,401 samples. 
Each sample strictly adheres to three criteria during the filtering process: it must rely on visual content comprehension, cover a broad range of ability dimensions, and require advanced multimodal reasoning capabilities. 
Therefore, using MMStar for evaluation can better reflect the capabilities of LVLMs. 

\textbf{CHAIR. }CHAIR~\cite{rohrbach2018object} has established two metrics, $CHAIR_S$ and $CHAIR_I$, to assess the degree of hallucination in the generated responses.
Where $CHAIR_S=\frac{|\{captions \ with \ hallucinated\ objects\}|}{|\{all \ captions\}|}$ indicates the degree of hallucination at the sentence level, while $CHAIR_I=\frac{|\{hallucinated\ objects\}|}{|\{all \ mentioned \ objects\}|}$ represents the degree of hallucination at the object level.
Following previous work, we randomly sampled 500 samples and used \textit{`Please describe this image in detail.'} to guide the LVLMs in generating captions for the images. 
\section{Result on InternVL2.5-MPO}
In order to further demonstrate the effectiveness of INTER, we conducted a comparison on the current state-of-the-art LVLM InternVL2.5-MPO (8B)~\cite{chen2024internvl}.  As shown in \cref{mpo}, the performance of INTER is superior to the baseline methods across various benchmarks. Moreover, `Nucleus+INTER' performs better than VCD~\cite{leng2023mitigating} across all benchmarks, while `Beam+INTER' also performs better than OPERA~\cite{huang2024opera}.

\begin{table*}[t!]
\tablestyle{4pt}{1.2}
\begin{tabular}{lccccccccc}
model& benchmark&Nucleus& Nucleus+INTER&Beam& Beam+INTER&VCD\textsuperscript{*}& VCD\textsuperscript{*}+INTER&OPERA\textsuperscript{\dag}\\\hline
\multirow{6}{*}{\makecell{InternVL2.5-MPO (8B)\\~\cite{liu2023improvedllava}}}    &MME (\textit{Total Score})~\cite{fu2023mme} $\uparrow$ &2175.7&\textbf{2204.8}	&2298.3	&\textbf{2316.4}&2189.2&\textbf{2209.9}&2299.7	\\
&  POPE (MSCOCO)~\cite{li2023pope} $\uparrow$&85.7&\textbf{89.2}	&88.7	&\textbf{89.3}&88.6&88.5&88.9\\
&  MM-Bench~\cite{liu2023mmbench} $\uparrow$&80.1&\textbf{81.5}	&84.4	&\textbf{84.6}&80.8&\textbf{81.6}&84.4\\
    &MMStar~\cite{chen2024we} $\uparrow$ &60.8&\textbf{62.5}	&63.0	&\textbf{63.9}&61.9&\textbf{63.3}&63.5	\\
    &  CHAIR ($C_S$+$C_I$)~\cite{rohrbach2018object} $\downarrow$&25.2&\textbf{21.6}	&23.6	&\textbf{19.7}&25.5&25.9&22.0\\
     &  LLaVA-Bench ~\cite{llava_bench_in_the_wild} $\uparrow$&9.5&\textbf{11.9}	&9.3	&\textbf{12.5}&10.1&\textbf{11.2}&10.5\\
    \hline
\end{tabular}
\caption{\textbf{Validation of INTER on the state-of-the-art LVLM InternVL2.5-MPO~\cite{chen2024internvl}.} \textsuperscript{*} and \textsuperscript{\dag} represent correction based on Nucleus Sampling and Beam Search. }
\label{mpo}
\end{table*}
\section{Ablation Study on  Interaction Guide Locator. }
In addition to the effectiveness analysis of the Interaction Guide Locator based on Beam Search~\cite{graves2012sequence, sutskever2014sequence,boulanger2013audio}, we also conducted ablation experiments on various decoding strategies for IGL. 
As shown in \cref{tab:ab1}, we evaluated the performance improvement brought by IGL on MME~\cite{fu2023mme}. 
It can be observed that the performance significantly decreases without IGL across all decoding strategies, suggesting that IGL identifies the positions of keywords, preventing the excessive guidance of interactions, thereby effectively improving performance. 

\begin{table}[!t]
\tablestyle{5pt}{1.2}
\begin{tabular}{lccc}
method&\makecell{InstructBLIP\\~\cite{instructblip}} & \makecell{LLaVA-v1.5\\~\cite{liu2023improvedllava}} &\makecell{mPLUG-owl2\\~\cite{ye2024mplug}}\\\hline
 \vb Nucleus+IPM &1569.7 	&1690.9	&	1640.6 \\
\gr \cb Nucleus+INTER&\textbf{1595.5}&\textbf{1731.6}&\textbf{1641.7}\\
 \vb Beam+IPM  &	1556.2 &1648.6 	&	1623.0 \\
\gr\cb Beam+INTER&\textbf{1562.2}&\textbf{1744.0}&\textbf{1716.1}\\
 \vb VCD\textsuperscript{*}+IPM &1583.6 	&1700.0&	1620.1\\
\gr\cb VCD\textsuperscript{*}+INTER &\textbf{1605.0}&\textbf{1749.6}&\textbf{1626.2}\\
\vb OPERA\textsuperscript{\dag}+IPM&1553.5 &1720.8 &1625.7 \\
\gr\cb OPERA\textsuperscript{\dag}+INTER&\textbf{1567.0} &\textbf{1727.4} &\textbf{1741.7} \\

\end{tabular}
\caption{\textbf{Ablation Study on Interaction Guide Locator (IGL).} }
\label{tab:ab1}
\end{table}

\section{Parameter Analysis of Interaction Guide Locator.}
\begin{figure}[!t]
    \centering
    \includegraphics[width=1.0\linewidth, trim = 0 0 0 7,clip]{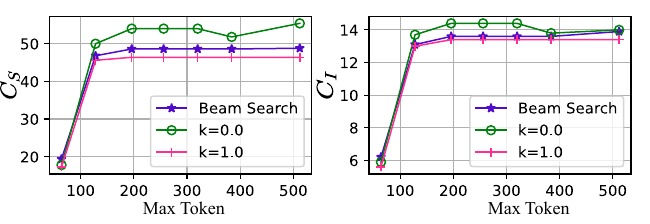}
    \caption{\textbf{Parameter analysis of $k$ in Interaction Guide Locator. }
    Evaluation of $C_I$ and $C_S$ after using different $k$ to guide Beam Search~\cite{graves2012sequence, sutskever2014sequence,boulanger2013audio} on various lengths on CHAIR~\cite{rohrbach2018object}. 
    }
    \label{fig:caption_k}
\end{figure}

Through experiments on CHAIR~\cite{rohrbach2018object} and MME~\cite{fu2023mme} benchmarks, we analyze how the interaction guidance coefficient $k$ affects the performance of INTER. 

As shown in \cref{fig:caption_k}, varying $k$ values lead to significantly different behaviors in LLaVA-v1.5. 
When $k=0.0$ which applies the Interaction Probability Modifier at all decoding steps, we observe reduced hallucination for short sequences after using INTER. 
However, this approach harms performance in longer sequences due to unnecessary modifications at non-critical positions, as evidenced by the performance drop compared to $k=1.0$. 

\begin{figure}[!t]
    \centering
\includegraphics[width=1.0\linewidth, trim = 0 0 0 7,clip]{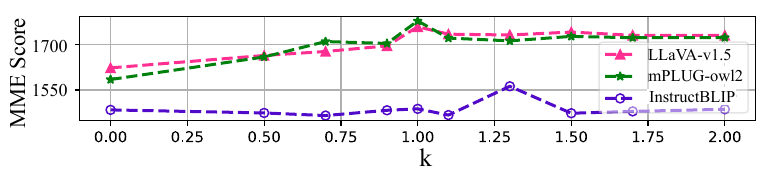}
    \caption{\textbf{Parameter analysis of $k$ on MME~\cite{fu2023mme}. }Each value represents total score of using INTER on Beam Search~\cite{graves2012sequence, sutskever2014sequence,boulanger2013audio}. }
    \label{fig:k}
\end{figure}

\cref{fig:k} reveals model-dependent optimal $k$ values. 
On MME, InstructBLIP achieves peak performance at $k=1.3$, beyond which excessive adjustment suppression causes gradual performance degradation. 
This suggests a balance between necessary corrections and interference avoidance. 

\section{Result on POPE}
\label{gqa}
\begin{table}[!t]
\tablestyle{1pt}{1.2}
\begin{tabular}{lcccc}
method&   
\makecell{InstructBLIP\\~\cite{instructblip}} & 
\makecell{LLaVA-v1.5\\(7B)~\cite{liu2023improvedllava}}& 
\makecell{Qwen-VL\\~\cite{bai2023qwen} }& 
\makecell{mPLUG-owl2\\~\cite{ye2024mplug}}  \\ \hline
\vb Nucleus~\cite{holtzman2019curious} &  77.0 & 79.1 & 76.1 & 76.8 \\
\gr
\cb Nucleus+INTER  & 
\textbf{81.9}  & 
\textbf{84.3}  & 
\textbf{81.9} & 
\textbf{80.2 }
\\
\vb Greedy~\cite{song2024good} &  81.3 & 85.1 & 79.4 & 80.9\\
\gr
\cb Greedy+INTER  & 
\textbf{82.2} & 
\textbf{85.2}  & 
\textbf{81.4}  & 
\textbf{80.9 }
\\
 \vb Beam~\cite{graves2012sequence, sutskever2014sequence,boulanger2013audio} & 81.4 & 84.7 & 79.7 &80.2  \\
\gr
 \cb Beam+INTER 
 &\textbf{83.3}
&\textbf{84.7}
 &\textbf{81.2}
 &\textbf{81.0}
  \\
 \vb VCD\textsuperscript{*}~\cite{leng2023mitigating} &80.6 & 82.7 & 82.3 &79.7 \\
\gr
\cb VCD\textsuperscript{*}+INTER & \textbf{80.9}
 & \textbf{83.6}
 & \textbf{82.0} 
 & \textbf{79.5} 
 \\
\vb OPERA\textsuperscript{\dag}~\cite{huang2024opera}  & 81.3&  84.9& 79.8 &  80.4\\
\gr
\cb OPERA\textsuperscript{\dag}+INTER & \textbf{82.5} 
 & \textbf{85.8}
 & \textbf{83.1}
 &\textbf{81.1} 
 \\
\end{tabular}
\caption{\textbf{Evaluating the performance of INTER's correction on four decoding strategies} by the mean F1-score across various partitions of GQA~\cite{hudson2019gqa}. Higher values are better.}
\label{tab:popegqa}
\end{table}
In this subsection, we evaluate the performance of the proposed INTER on the GQA~\cite{hudson2019gqa} dataset within the POPE benchmark. The results, as shown in the \cref{tab:popegqa}, indicate that significant performance improvements across four models. Furthermore, these enhancements are consistent with the results on the MSCOCO~\cite{lin2014microsoft} and AOKVQA~\cite{schwenk2022okvqa} datasets, further validating the effectiveness and robustness of our approach. 

\section{Result on MME}
In addition to demonstrating the performance improvements brought by INTER across various decoding strategies in 14 subtasks, we also conducted comparisons in terms of the total score and perception total score of MME~\cite{fu2023mme}. 
As shown in \cref{tab:mme1}, after correction with INTER, there was a maximum increase of over 343.7 points in the total score compared to Nucleus Sampling, and a maximum increase of over 311.2 points in the perception total score. Furthermore, it can be observed that there is a certain degree of improvement across different models and decoding strategies, indicating the effectiveness of INTER. 
\begin{table*}[!t]
\tablestyle{2pt}{1.05}
\begin{tabular}{lcccccccc}
\multirow{2}{*}{method}&   \multicolumn{2}{c}{InstructBLIP~\cite{instructblip}} & \multicolumn{2}{c}{LLaVA-v1.5 (7B)~\cite{liu2023improvedllava}}& \multicolumn{2}{c}{Qwen-VL~\cite{bai2023qwen}} & \multicolumn{2}{c}{mPLUG-owl2~\cite{ye2024mplug}}  \\
&\textit{\makecell{Perception \\Total}}&\textit{Total}
&\textit{\makecell{Perception \\Total}}&\textit{Total}
&\textit{\makecell{Perception \\Total}}&\textit{Total}
&\textit{\makecell{Perception \\Total}}&\textit{Total} \\ 
  \hline
\vb 
Nucleus~\cite{holtzman2019curious}& 984.4& 1251.8 & 1279.2&1502.2&1216.6& 1465.6&1266.3& 1573.5\\
\gr
\cb Nucleus+INTER  
&1295.6 ($\uparrow$311.2)
&1595.5 ($\uparrow$343.7) 
&1372.0 ($\uparrow$92.8) 
&1731.6 ($\uparrow$229.4) 
&1279.8 ($\uparrow$63.2)
&1542.9 ($\uparrow$77.3)
&1353.8($\uparrow$87.5)
&1641.7 ($\uparrow$68.2)
\\
 \vb Greedy~\cite{song2024good} & 1160.9& 1419.8 & 1452.2 &1750.4 &1238.9&1512.5&1352.5& 1709.3  \\
\gr
 \cb Greedy+INTER 
 & 1291.2 ($\uparrow$130.3)
 &1593.3 ($\uparrow$173.5)
 &1470.5 ($\uparrow$18.3)
 &1761.3 ($\uparrow$10.9)
 &1292.4 ($\uparrow$53.5)
 &1544.2 ($\uparrow$31.7)
 &1360.4 ($\uparrow$7.9)
 & 1731.4 ($\uparrow$22.1) \\
 \vb Beam~\cite{graves2012sequence, sutskever2014sequence,boulanger2013audio} & 1128.9& 1318.6 & 1409.4 &1707.2 &1229.5&1513.4&1358.4& 1710.5  \\
\gr
 \cb Beam+INTER 
 & 1281.8 ($\uparrow$152.9)
 &1562.2 ($\uparrow$243.6)
 &1438.3 ($\uparrow$28.9)
 &1744.0 ($\uparrow$36.8)
 &1271.5 ($\uparrow$42.0)
 &1575.0 ($\uparrow$61.6)
 &1363.3 ($\uparrow$4.9)
 & 1716.1 ($\uparrow$5.6) \\
 \vb VCD\textsuperscript{*}~\cite{leng2023mitigating} 
 & 1167.9&1487.1
 & 1364.0& 1716.0 
 &1240.0&1546.5
&  1269.7&1573.0\\
\gr
\cb VCD\textsuperscript{*}+INTER 
&1306.8 ($\uparrow$138.9)
&1605.0 ($\uparrow$117.9) 
&1380.0 ($\uparrow$16.0) 
&1749.6 ($\uparrow$33.6)
&1297.0 ($\uparrow$57.0) 
&1575.6 ($\uparrow$29.1)
&1305.2 ($\uparrow$35.5) 
&1626.2 ($\uparrow$53.2) \\
\vb OPERA\textsuperscript{\dag}~\cite{huang2024opera}  
& 1137.5& 1326.5& 1430.8 & 1721.2 &1228.8&1501.7& 1357.6& 1740.8  \\
\gr
\cb OPERA\textsuperscript{\dag}+INTER 
& 1274.1 ($\uparrow$136.6) 
&1567.0 ($\uparrow$240.5) 
&1439.8 ($\uparrow$9.0) 
&1727.4 ($\uparrow$6.2)
& 1304.0 ($\uparrow$75.2) 
&1564.4 ($\uparrow$62.7)&
1377.7 ($\uparrow$20.1) & 
1741.7 ($\uparrow$0.9) \\
\end{tabular}
\caption{\textbf{The total scores and perceptual total scores on MME~\cite{fu2023mme}.} 
$\uparrow$ means that higher values indicate lower hallucination levels. 
Results showed that the addition of INTER led to a certain degree of mitigating hallucinations for all deocding strategies. }
\label{tab:mme1}
\end{table*}

\section{Result on MM-Bench}
\label{mmbench}
\begin{table}[t!]
\tablestyle{4pt}{1.2}
\begin{tabular}{lccccccc}
 method&Overall& AR&CP  &FP-C&FP-S&LR&RR\\\hline
 \vb Nucleus~\cite{holtzman2019curious} &57.3&50.0	&73.0	&48.8&58.1&44.7&52.6	\\
 \gr \cb Nucleus+INTER &\textbf{62.6}&\textbf{57.0}	&\textbf{76.4}	&\textbf{52.0}&\textbf{65.0}&\textbf{48.3}&\textbf{63.2}	\\
     \vb Greedy~\cite{song2024good} &65.2&58.4&77.8&55.3&69.3&50.7&64.7\\
\gr \cb Greedy+INTER &\textbf{65.2}&\textbf{58.4}&\textbf{77.8}&\textbf{55.3}&\textbf{69.3}&\textbf{50.7}&\textbf{64.7}
\\
   \vb Beam~\cite{graves2012sequence, sutskever2014sequence,boulanger2013audio} &65.1&57.8&77.8&55.3&69.3&50.7&64.7\\
\gr \cb Beam+INTER &\textbf{65.1}&\textbf{58.1}&\textbf{77.8}&\textbf{55.3}&\textbf{69.3}&\textbf{50.7}&\textbf{64.7}
\\
\vb VCD\textsuperscript{*} ~\cite{leng2023mitigating}&62.5&54.7&77.5&53.4&64.9&48.5&60.5\\
\gr \cb VCD\textsuperscript{*}+INTER&\textbf{62.9}&\textbf{54.7}&\textbf{77.6}&52.3&\textbf{66.0}&47.6&\textbf{65.0} \\
\vb 
OPERA\textsuperscript{\dag}~\cite{huang2024opera}&65.0&57.8&77.8&55.3&69.0&50.7&64.7 \\
\gr \cb OPERA\textsuperscript{\dag}+INTER&\textbf{65.0}&\textbf{57.8}&\textbf{77.9}&55.1&\textbf{69.0}&\textbf{50.7}&\textbf{64.7} \\
\end{tabular}
\caption{\textbf{Validation of INTER on MM-Bench~\cite{liu2023mmbench} using LLaVA-v1.5 (7B)~\cite{liu2023improvedllava}. }}
\label{tab:mmbench}
\end{table}
\begin{table}[t!]
\tablestyle{4pt}{1.2}
\begin{tabular}{lccccccc}
 method&Overall& AR&CP  &FP-C&FP-S&LR&RR\\\hline
 \vb Nucleus~\cite{holtzman2019curious} &57.0&51.1	&72.2	&41.5&59.9&44.6&55.3	\\
 \gr \cb Nucleus+INTER &\textbf{61.4}&\textbf{57.6}	&\textbf{74.4}	&\textbf{41.7}&\textbf{62.8}&\textbf{46.9}&\textbf{60.8}	\\
    \vb Greedy~\cite{song2024good} &63.5&55.4&77.5&52.2&65.8&49.1&70.7\\
\gr \cb Greedy+INTER &62.3&\textbf{56.3}&75.7&49.3&65.3&\textbf{49.3}&64.7\\
   \vb Beam ~\cite{graves2012sequence, sutskever2014sequence,boulanger2013audio} &63.5&55.4&77.5&52.2&65.8&49.8&69.2\\
\gr \cb Beam+INTER &\textbf{63.7}&\textbf{55.6}&\textbf{77.5}&\textbf{52.4}&\textbf{65.9}&\textbf{49.8}&\textbf{70.7}\\
\vb VCD\textsuperscript{*}~\cite{leng2023mitigating}  &59.2&52.1&75.1&43.7&62.1&45.9&59.0\\
\gr \cb VCD\textsuperscript{*}+INTER&\textbf{59.5}&\textbf{53.6}&74.1&\textbf{44.3}&\textbf{62.4}&\textbf{48.0}&58.3 \\
\vb 
OPERA\textsuperscript{\dag}~\cite{huang2024opera}&63.4&55.4&77.5&52.2&65.8&49.1&69.2 \\
\gr \cb OPERA\textsuperscript{\dag}+INTER&\textbf{63.6}&\textbf{55.4}&\textbf{77.5}&\textbf{52.3}&65.7&\textbf{49.8}&\textbf{70.9}\\
\end{tabular}
\caption{\textbf{Validation of INTER on MM-Bench~\cite{liu2023mmbench} using mPLUG-owl2~\cite{ye2024mplug}. }}
\label{tab:mmbench-mplugowl2}
\end{table}
\begin{table}[t!]
\tablestyle{4.5pt}{1.2}
\begin{tabular}{lccccccc}
 method&Avg.& CP&FP  &IR&LR&ST&MA\\\hline
 \vb Nucleus~\cite{holtzman2019curious}  &29.3&52.8	&22.4	&38.0&22.8&17.6&22.4	\\
 \gr \cb Nucleus+INTER &\textbf{31.9}&\textbf{58.0}	&\textbf{29.8}	&\textbf{39.4}&\textbf{27.6}&15.4&\textbf{22.8}	\\
 \vb Greedy~\cite{song2024good} &30.7&59.2&24.8&40.0&27.2&13.6&19.6\\
\gr  \cb Greedy+INTER 
&\textbf{31.9}&55.2	&\textbf{29.2}	&\textbf{45.2}&\textbf{29.2}&\textbf{15.2}&17.6\\
   \vb Beam~\cite{graves2012sequence, sutskever2014sequence,boulanger2013audio} &31.1&58.4&22.8&40.4&28.8&14.8&21.2\\
\gr \cb Beam+INTER 
&\textbf{31.7}&54.8	&\textbf{30.4}	&\textbf{44.0}&26.0&\textbf{18.4}&16.4\\
\vb VCD\textsuperscript{*}~\cite{leng2023mitigating} &30.3&54.4&24.4&38.0&26.0&16.8&22.4\\
\gr \cb VCD\textsuperscript{*}+INTER&\textbf{31.1}&\textbf{55.6}&\textbf{26.8}&\textbf{40.0}&\textbf{28.8}&15.2&20.4 \\
\vb OPERA\textsuperscript{\dag}~\cite{huang2024opera} &31.4&59.2&23.6&40.8&28.8&14.8&21.2\\
\gr \cb OPERA\textsuperscript{\dag}+INTER&\textbf{32.9}&56.8&\textbf{30.0}&\textbf{42.4}&\textbf{28.8}&\textbf{18.8}&20.8 \\
\end{tabular}
\caption{\textbf{Validation of INTER on MMStar~\cite{chen2024we} using LLaVA-v1.5 (7B)~\cite{liu2023improvedllava}. }}
\label{tab:MMStar}
\end{table}

\begin{table}[t!]
\tablestyle{5pt}{1.2}
\begin{tabular}{lccccccc}
 method&Avg.& CP&FP  &IR&LR&ST&MA\\\hline
 \vb Nucleus~\cite{holtzman2019curious} &30.5&50.4	&24.8	&40.0&27.6&17.6&22.4	\\
 \gr \cb Nucleus+INTER &\textbf{32.3}&\textbf{53.2}	&\textbf{25.2}	&\textbf{42.0}&\textbf{31.6}&\textbf{18.0}&\textbf{23.6}	\\
 \vb Greedy~\cite{song2024good} &30.1&53.2&26.0&40.4&28.8&12.0&20.0\\
\gr\cb Greedy+INTER 
&\textbf{33.3}&\textbf{55.2}	&\textbf{29.6}	&\textbf{42.4}&\textbf{30.8}&\textbf{17.2}&\textbf{24.8}\\
   \vb Beam~\cite{graves2012sequence, sutskever2014sequence,boulanger2013audio} &30.7&52.4&26.0&42.0&28.8&12.0&23.2\\
\gr \cb Beam+INTER 
&30.5&\textbf{52.8}	&24.4&39.2&\textbf{29.6}&\textbf{14.0}&22.8\\
\vb VCD\textsuperscript{*}~\cite{leng2023mitigating} &31.5&52.8&26.8&36.0&25.6&18.8&28.8\\
\gr \cb VCD\textsuperscript{*}+INTER&\textbf{32.3}&\textbf{52.8}&25.6&\textbf{39.6}&\textbf{32.0}&17.2&26.8 \\
\vb OPERA\textsuperscript{\dag}~\cite{huang2024opera} &30.5&52.8&26.4&41.6&28.4&12.0&22.0\\
\gr \cb OPERA\textsuperscript{\dag}+INTER&\textbf{30.9}&\textbf{52.8}&24.4&39.6&\textbf{29.6}&\textbf{14.8}&\textbf{24.0} \\

\end{tabular}
\caption{\textbf{Validation of INTER on MMStar~\cite{chen2024we} using mPLUG-owl2~\cite{ye2024mplug}.}}
\label{tab:MMStar-mplug}
\end{table}

To illustrate the improvement of INTER on MM-Bench in more detail, we present the performance of each subtask in \cref{tab:mmbench,tab:mmbench-llava-l3-p,tab:mmbench-llava-l3-r}. 
As we can see, using INTER results in an improvement across various metrics. 
In addition, to validate the performance of INTER across different LVLMs, \cref{tab:mmbench-mplugowl2} presents the performance on mPLUG-owl2. It can be observed that there is a high consistency with LLaVA-v1.5, and INTER brings a certain degree of enhancement. 
Finally, detailed results of mPLUG-owl2 at each subtasks are also presented in \cref{tab:mmbench-mplug-l3-p,tab:mmbench-mplug-l3-r}.

\section{Result on MMStar}
\label{MMStar}

Likewise, to assess the effectiveness of INTER on MMStar, we also present the performance of each subtask on LLaVA-v1.5 (7B)~\cite{liu2023improvedllava} in \cref{tab:MMStar-llava-l3-2,tab:MMStar-llava-l3-p,tab:MMStar}. 
The results indicate that our approach achieves good performance across most subtasks. 
Although there is no improvement of the correction effects on VCD~\cite{leng2023mitigating} and OPERA~\cite{huang2024opera} in the `Math', the correction results using INTER for `Nucleus' outperform those of VCD, and the performance on Beam Search is better than OPERA. 

In addition, we conducted comparative experiments on MMStar using mPLUG-owl2 in \cref{tab:MMStar-mplug-l3-1,tab:MMStar-mplug-l3-2,tab:MMStar-mplug}, and the results show that our method has a certain corrective effect across different LVLMs. 

\section{Result on Greedy Search}
\label{greedy}
\begin{table}[!t]
\centering
\tablestyle{1pt}{1.0}
\begin{tabular}{c|c|cccc}
\hline
\multirow{2}{*}{model} & \multirow{2}{*}{method}&\multicolumn{2}{c}{CHAIR (512)} & \multirow{2}{*}{MMStar$\uparrow$} & \multirow{2}{*}{MME$\uparrow$} \\
& &$C_s$$\downarrow$ & $C_I$$\downarrow$ & & \\
\hline
\multirow{4}{*}{\textbf{InstructBLIP~\cite{instructblip}}} &
M3ID~\cite{favero2024multi} & 63.1 & 21.1 & 29.8 & 1440.6 \\
&Ritual~\cite{woo2024ritual} & 62.1 & 20.9 & 29.5 & 1576.7 \\
&SID~\cite{huo2024self} & 59.7 & 21.4 & 28.1 & 1385.1 \\
&\textbf{INTER (ours)} & \textbf{59.0} & \textbf{20.8} & \textbf{30.5} & \textbf{1595.5} \\
\hline

\multirow{4}{*}{\textbf{LLaVA-v1.5~\cite{liu2023improvedllava}}} &
M3ID~\cite{favero2024multi} & 67.1 & 19.7 & 30.9 & 1322.9 \\
&Ritual~\cite{woo2024ritual} & 52.4 & 15.8 & 31.2 & 1754.7 \\
&SID ~\cite{huo2024self}& 52.0 & 14.3 & 31.0 & 1692.4 \\
&\textbf{INTER (ours)} & \textbf{51.8} & \textbf{14.1} & \textbf{31.9} & 1731.6 \\
\hline
\end{tabular}
\caption{Comparison with other methods. }
\label{tab:R1-1}
\end{table}
\begin{table*}[!t]
\tablestyle{5pt}{1.2}
\begin{tabular}{lcccccccc}
\multirow{2}{*}{method}&   
\multicolumn{2}{c}{InstructBLIP~\cite{instructblip}} & 
\multicolumn{2}{c}{LLaVA-v1.5 (7B)~\cite{liu2023improvedllava}}& 
\multicolumn{2}{c}{Qwen-VL~\cite{bai2023qwen}} & 
\multicolumn{2}{c}{mPLUG-owl2~\cite{ye2024mplug}}  \\
& COCO~\cite{lin2014microsoft} & AOKVQA~\cite{schwenk2022okvqa} & COCO~\cite{lin2014microsoft} & AOKVQA~\cite{schwenk2022okvqa}& COCO~\cite{lin2014microsoft} & AOKVQA~\cite{schwenk2022okvqa}  & COCO~\cite{lin2014microsoft} & AOKVQA~\cite{schwenk2022okvqa}\\ \hline
 \vb Greedy~\cite{song2024good} 
 &  84.4
 &  81.3
 & 84.5 
 & 84.3
 & 83.4
 & 	85.0
 & 83.4
 &  81.3 \\
\gr
 \cb Greedy+INTER 
 & \textbf{85.1}
 &  \textbf{83.7 }
 & \textbf{86.4 }
  & \textbf{84.4 }
& \textbf{86.0 }
&\textbf{86.4}
 & \textbf{83.4 }
& \textbf{81.7 }
  \\
\end{tabular}
\caption{\textbf{Evaluating the performance of INTER's correction on Greedy Search~\cite{song2024good}} by the mean F1-score across various partitions of two datasets in POPE~\cite{li2023pope}. Higher values are better. }
\label{tab:pope1_greedy}
\end{table*}
\begin{table}[!t]
\tablestyle{2pt}{1.2}
\begin{tabular}{llcccccc}
\multirow{2}{*}{\makecell{Max \\Token}}& \multirow{2}{*}{method} & \multicolumn{2}{c}{\makecell{InstructBLIP\\~\cite{instructblip}}} & \multicolumn{2}{c}{\makecell{LLaVA-v1.5 (7B)\\~\cite{liu2023improvedllava}}} &\multicolumn{2}{c}{\makecell{mPLUG-owl2\\~\cite{ye2024mplug}}}\\
&&$C_S$&$C_I$&$C_S$&$C_I$&$C_S$&$C_I$ \\\hline
& \vb Greedy~\cite{song2024good} & 26.2&	13.8&22.0&	6.7&	23.0&	8.3\\
\gr64&\cb Greedy+INTER &\textbf{25.8}	&\textbf{9.2}&	\textbf{22.0}&	\textbf{6.7}	&\textbf{20.6}&	\textbf{7.9}\\\hline
 &\vb Greedy~\cite{song2024good} & 49.2	&21.9	&48.8&	13.4	&58.2	&18.5\\
\gr512&\cb Greedy+INTER &55.8	&\textbf{18.1}	&\textbf{48.8}	&\textbf{13.4}	&\textbf{54.4}&	\textbf{17.9}\\

\end{tabular}
\caption{\textbf{Evaluating the effectiveness of INTER in correcting Greedy Search} using LLaVA-v1.5 on CHAIR~\cite{rohrbach2018object}, with a maximum token length of 64 and 512. A smaller value indicates a lower degree of hallucinations. }
\label{tab:chair64_greedy}
\end{table}
\begin{figure*}[!t]
    \centering
\includegraphics[width=1.0\linewidth]{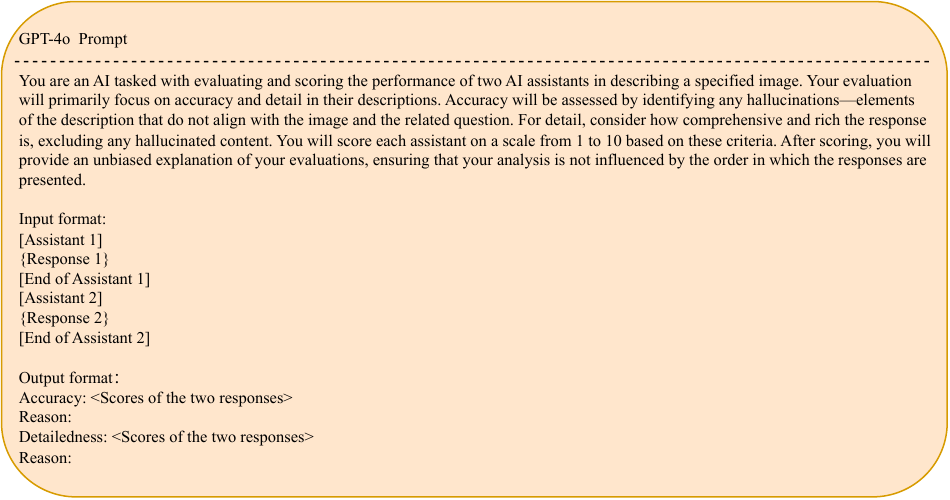}
    \caption{\textbf{Prompts of GPT-4o~\cite{hurst2024gpt} for evaluations.} }
    \label{fig:gpt4pprompt}
\end{figure*}

\begin{figure}[!t]
    \centering
\includegraphics[width=1.0\linewidth]{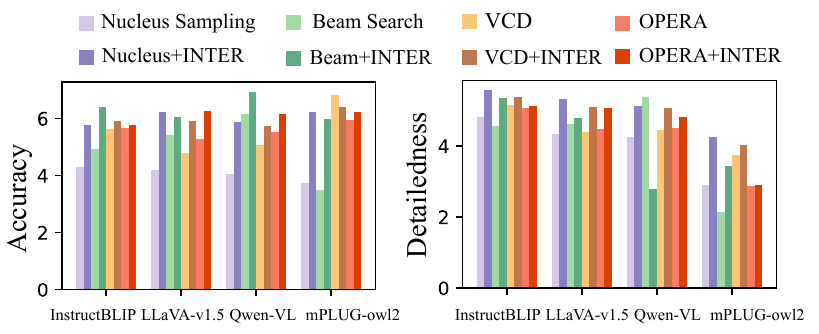}
    \caption{\textbf{Evaluating accuracy and detailedness on response of 60 Image-Text pairs in LLaVA-Bench~\cite{llava_bench_in_the_wild}} using GPT-4o~\cite{hurst2024gpt}. }
    \label{fig:llava-bench-quanti}
\end{figure}

In \cref{tab:pope1_greedy,tab:mme1,tab:popegqa,tab:chair64_greedy,tab:MMStar-mplug-l3-1,tab:MMStar-mplug-l3-2,tab:MMStar-mplug,tab:MMStar-llava-l3-2,tab:MMStar-llava-l3-p,tab:MMStar,tab:mmbench,tab:mmbench-llava-l3-p,tab:mmbench-llava-l3-r,tab:mmbench-mplugowl2,tab:mmbench-mplug-l3-p,tab:mmbench-mplug-l3-r}, we demonstrated the effectiveness of INTER in correcting the Greedy Search across various benchmarks. 
It is evident that there is a significant improvement across different benchmarks, indicating that our method INTER exhibits generalization capabilities in correcting various decoding strategies.

\section{Result on LLaVA-Bench}

\begin{figure*}[!t]
    \centering
    \includegraphics[width=0.85\linewidth]{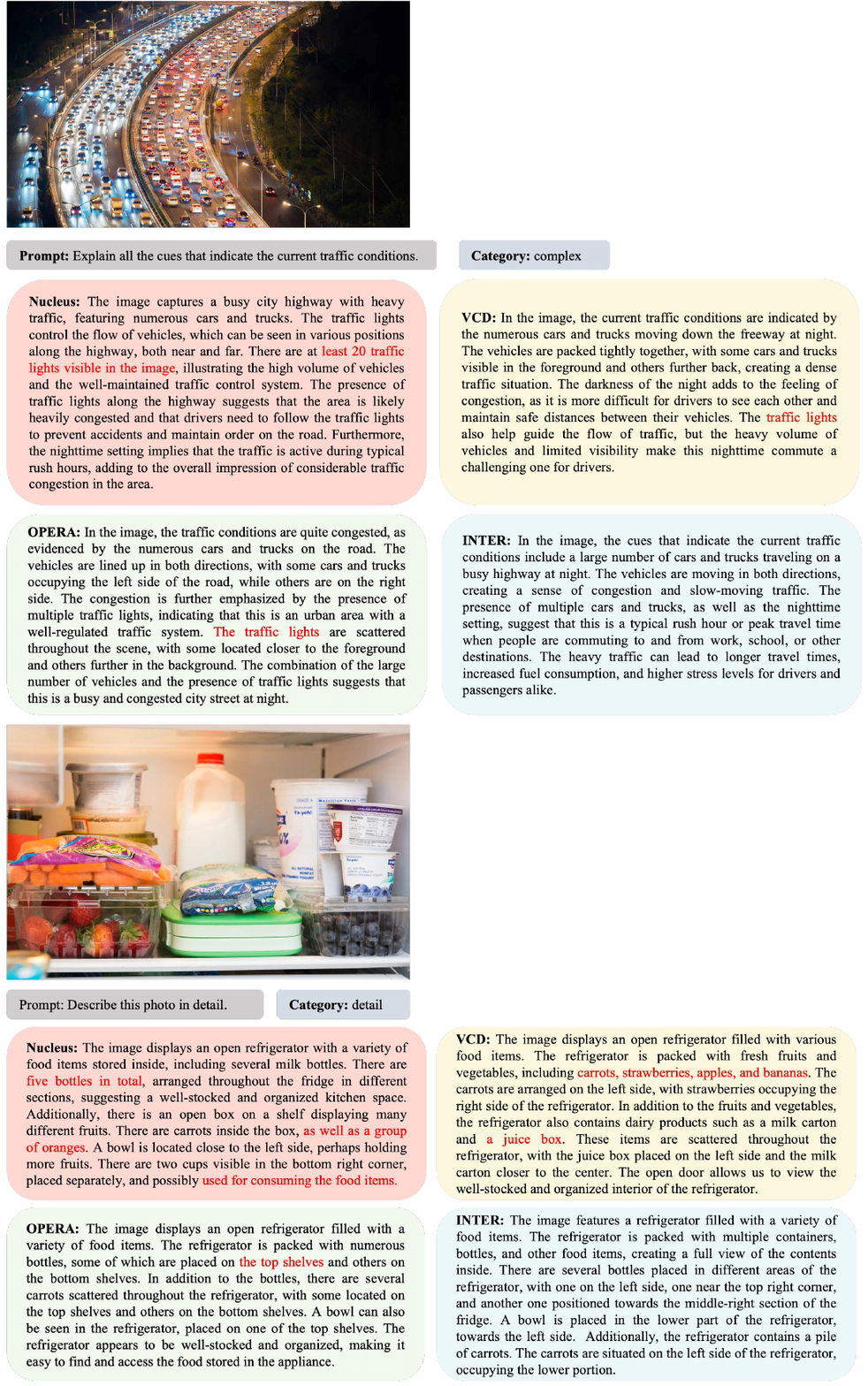}
    \caption{\textbf{Case study on LLaVA-v1.5} through LLaVA-Bench. Hallucinations are marked in red.}
    \label{fig:vis_llava}
\end{figure*}

\begin{figure*}[!t]
        \centering
    \includegraphics[width=0.85\linewidth]{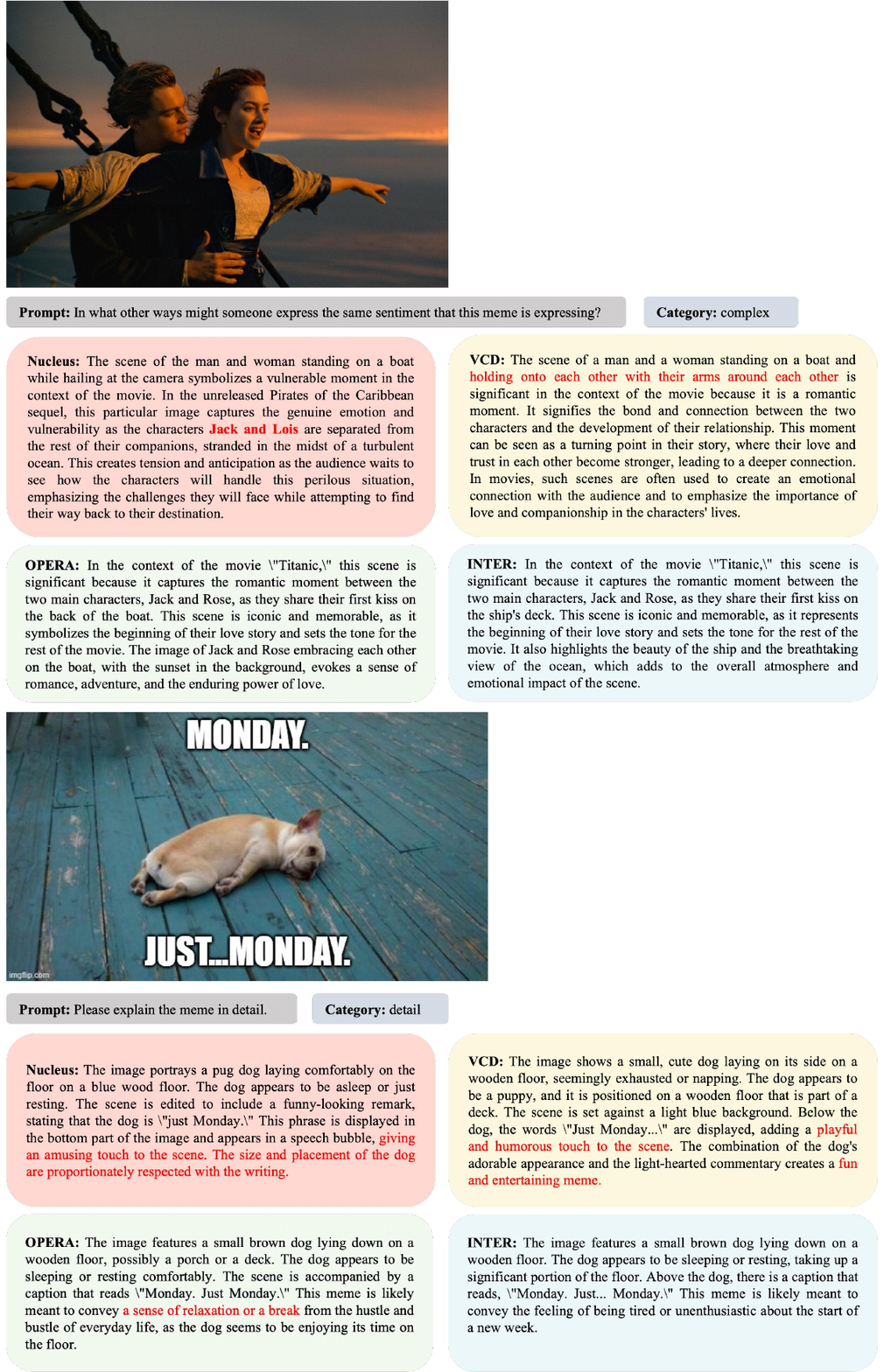}
    \caption{\textbf{Case study on mPLUG-owl2} through LLaVA-Bench. Hallucinations are marked in red. }
    \label{fig:vis_mplug}
\end{figure*}

\begin{figure*}[!t]
    \centering
    \includegraphics[width=0.85\linewidth]{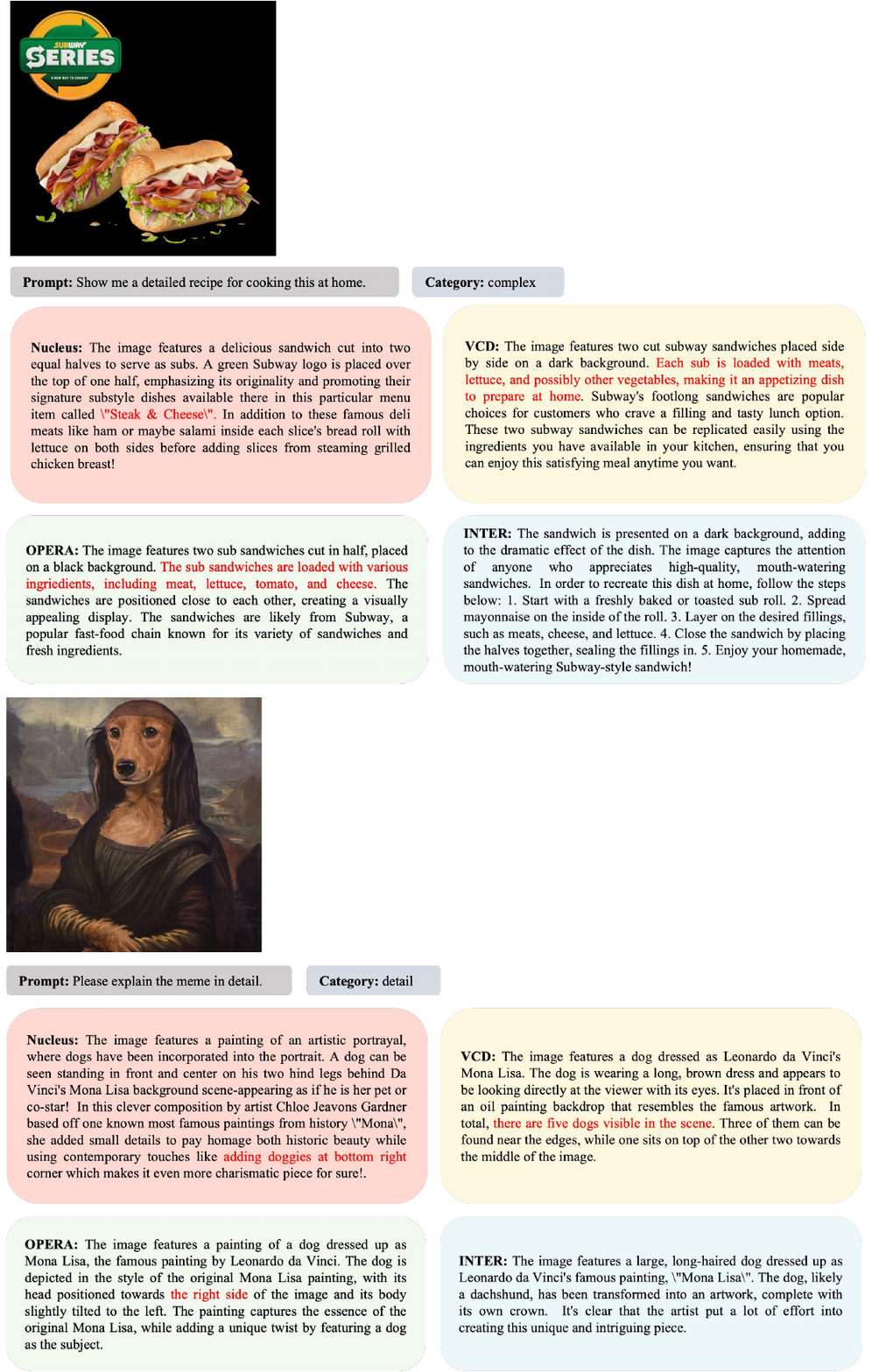}
    \caption{\textbf{Case study on InstructBLIP} through LLaVA-Bench. Hallucinations are marked in red. }
    \label{fig:vis_blip}
\end{figure*}

To more intuitively demonstrate the performance of INTER, detailed case studies were conducted using LLaVA-Bench. In \cref{fig:vis_llava,fig:vis_mplug,fig:vis_blip}, examples of the captioning and complex reasoning task for each model are presented. The hallucination parts are highlighted in red.

In addition to case study, we also assessed the accuracy and detailedness of responses generated by various methods on the LLaVA-Bench dataset using GPT-4o~\cite{hurst2024gpt}. 
As shown in the Fig.~\ref{fig:llava-bench-quanti}, the answers generated after applying INTER calibration received higher scores. 
The template of prompt is shown in \cref{fig:gpt4pprompt}.

\section{Computation Efficiency}
Similar to VCD~\cite{leng2023mitigating}, which require additional forward passes, INTER also necessitates extra inference to compute the logits under different subsets of $A$. While INTER increases the total number of forward passes, the actual runtime overhead remains negligible due to the capability of compressing all subset evaluations into a single batch.

\section{Comparison with Other Methods. }
We conducted experiments with M3ID~\cite{favero2024multi}, Ritual~\cite{woo2024ritual} and SID~\cite{huo2024self} in \cref{tab:R1-1}. 
The results demonstrate that our INTER achieves comparable performance among compared methods. 

\section{Performance on Other Types of Tasks or Different LVLMs. }
We conducted experiments with DeepSeek-VL2~\cite{wu2024deepseek} on the visual grounding task. 
As shown in \cref{tab:deepseek-grounding}, results show that the INTER boosts the model performance on this task. 
 \begin{table}
 \tablestyle{8pt}{1.0}
 \begin{tabular}{ c c c }
 \hline
 \multirow{2}{*}{model}
 & \multicolumn{2}{c}{RefCOCO~\cite{yu2016modeling}}\\
 &testA &testB\\
 \hline
  DeepSeek-VL2-Tiny~\cite{wu2024deepseek}& 87.8 &  78.4  \\
  \gr +INTER& \textbf{88.6} &  \textbf{78.6} \\
 \end{tabular}
\caption{Performance on other types of tasks or different LVLMs. }
 \label{tab:deepseek-grounding}
 \end{table}

\section{The Range of the Harsanyi dividend. } 
The value range of $\mathrm{I}(A)^{\textbf{y}_{t}}$ could be influenced by several factors, \textit{e.g.}, benchmarks, LVLMs, etc. 
These complexities make it challenging to establish a theoretical bound for its value range. 
Nevertheless, we conducted experiments to empirically assess the distribution of $\mathrm{I}(A)^{\textbf{y}_{t}}$  in \cref{tab:range-of-value}. 
Moreover, when $\mathrm{I}(A)^{\textbf{y}_{t}}$ is negative, we consider that such interaction effects may hinder sampling this candidate token, which is considered similarly in prior studies~\cite{wang2020unified,zhang2021interpreting}.

\begin{table*}
 \tablestyle{0.5pt}{1}
 \begin{tabular}{l c c ccccc c c }
\multirow{3}{*}{Datasets}
 & \multirow{3}{*}{Order}
 & \multicolumn{2}{c}{\makecell{InstructBLIP~\cite{instructblip}}}
 & \multicolumn{2}{c}{\makecell{LLaVA-v1.5~\cite{liu2023improvedllava}} }
 & \multicolumn{2}{c}{\makecell{Qwen-VL~\cite{bai2023qwen}} }
 & \multicolumn{2}{c}{\makecell{mPLUG-owl2~\cite{ye2023mplugowl}}} \\
 & &\makecell{Mean \\Absolute Value}
 & Range&\makecell{Mean \\Absolute Value}  & Range&\makecell{Mean \\Absolute Value}&Range&\makecell{Mean \\Absolute Value}&Range \\
 \hline
 \multirow{2}{*}{MME~\cite{fu2023mme}}  & $\mathrm{I}\left(A\right|\{\textbf{v},\textbf{p}\})^{\textbf{y}_t}$ 
     & 0.80& [-9.8, 5.7] & 0.59 &[-13.8, 5.3]& 2.10& [-16.1, 10.0] & 0.07 &[-20.4, 27.8] \\
   & $\mathrm{I}\left(A\right|\{\textbf{p},\textbf{v}\})^{\textbf{y}_t}$ 
     & 4.07 &[-7.4, 22.8] & 0.59 &[-13.8, 9.3] & 1.93 &[-10.6, 8.2] & 0.07 &[-16.5, 10.2] \\
 \hline
 \multirow{2}{*}{CHAIR~\cite{rohrbach2018object}} & $\mathrm{I}\left(A\right|\{\textbf{v},\textbf{p}\})^{\textbf{y}_t}$ 
     & 0.60 &[-16.5, 14.2] & 0.06& [-3.3, 3.9] & 0.56& [-17.8, 12.4] & 0.16 &[-8.0, 7.7] \\
  & $\mathrm{I}\left(A\right|\{\textbf{p},\textbf{v}\})^{\textbf{y}_t}$ 
     & 0.60 &[-3.3, 3.9]& 0.06 &[-3.3, 3.9] & 0.94 &[-25.4, 8.1] & 0.16 &[-7.9, 7.5] \\
 \hline
 \end{tabular}
 \caption{The range of the metric $\mathrm{I}(A)^{\textbf{y}_{t}}$. }
 \label{tab:range-of-value}
 \end{table*}
 
\begin{table*}[ht!]
\tablestyle{2pt}{1.2}
\begin{tabular}{lcccccccccccc}
\hline
 &\multicolumn{5}{c}{Coarse Perception (CP)}  
 &\multicolumn{3}{c}{Cross-instance Fine-grained Perception (FP-C)}
 &\multicolumn{4}{c}{Single-instance Fine-grained Perception (FP-S)}\\ 
 method&\makecell{Image \\Emotion}
 &\makecell{Image \\Topic}
 &\makecell{Image \\Scene}
 &\makecell{Image \\Style}
 &\makecell{Image \\Quality}
 &\makecell{Action \\Recognition}
 &\makecell{Attribute \\Comparision}
 &\makecell{Spatial \\Realtionship}
 &\makecell{Celebrity \\Recognition}
 &\makecell{Object \\Localization}
 &\makecell{Attribute \\Recognition}
 &OCR\\\hline
 \vb Nucleus~\cite{holtzman2019curious} &71.0&76.4	&94.1	&66.5&24.7&82.3&44.0&11.9&75.5&25.7&73.5&53.2	\\
 \gr \cb Nucleus+INTER&\textbf{77.5}&\textbf{80.7}&\textbf{96.1}&\textbf{70.8}&\textbf{25.3}&\textbf{84.7}&\textbf{51.1}&
 \textbf{13.0}&\textbf{80.1}&\textbf{35.2}&\textbf{81.4}&\textbf{59.0} \\
    \vb Greedy~\cite{song2024good}&78.0&81.4&96.1&75.5&28.0&87.0&53.2&18.6&81.8&44.8&86.0&59.0\\  
\gr \cb Greedy+INTER &\textbf{78.0}&\textbf{81.4}&\textbf{96.1}&\textbf{75.5}&\textbf{28.0}&\textbf{87.0}&\textbf{53.2}&\textbf{18.6}&\textbf{81.8}&\textbf{44.8}&\textbf{86.0}&\textbf{59.0}
\\
   \vb Beam~\cite{graves2012sequence, sutskever2014sequence,boulanger2013audio}&78.0&81.4&96.1&75.5&28.0&87.0&53.2&18.6&81.8&44.8&86.0&59.0\\  
\gr \cb Beam+INTER &\textbf{78.0}&\textbf{81.4}&\textbf{96.1}&\textbf{75.5}&\textbf{28.0}&\textbf{87.0}&\textbf{53.2}&\textbf{18.6}&\textbf{81.8}&\textbf{44.8}&\textbf{86.0}&\textbf{59.0}
\\
\vb VCD\textsuperscript{*}~\cite{leng2023mitigating}&77.0&81.4&96.1&77.0&25.3&86.1&49.0&17.5&78.3&37.1&81.8&58.3\\
\gr \cb VCD\textsuperscript{*}+INTER
&\textbf{77.0} & \textbf{82.1} & \textbf{96.1} & 75.0 & \textbf{28.0} & 84.7 & 48.9 & 15.8 & \textbf{80.3} & \textbf{37.8} & 81.1 & \textbf{60.9}\\
\vb 
OPERA\textsuperscript{\dag}~\cite{huang2024opera} &78.0&81.4&96.1&75.5&28.0&87.0&53.2&18.6&81.8&43.5&86.0&59.0 \\
\gr \cb OPERA\textsuperscript{\dag}+INTER &\textbf{78.0}&\textbf{81.4}&\textbf{96.1}&\textbf{75.9}&\textbf{28.0}&\textbf{87.0}&\textbf{53.2}&17.9&\textbf{81.8}&\textbf{43.5}&\textbf{86.0}&\textbf{59.0}\\
\end{tabular}
\caption{\textbf{Evaluating the performance of INTER on MM-Bench~\cite{liu2023mmbench} using LLaVA-v1.5 (7B)~\cite{liu2023improvedllava}}, focusing on coarse perception and fine-grained perception subtasks.}
\label{tab:mmbench-llava-l3-p}
\end{table*}
\begin{table*}[ht!]
\tablestyle{4pt}{1.2}
\begin{tabular}{lccccccccccccccccccccc}
\hline
&
  \multicolumn{3}{c}{Attribute Reasoning (AR)}
 &\multicolumn{3}{c}{Logic Reasoning (LR)}
 &\multicolumn{2}{c}{Relation Reasoning (RR)}\\
 method&\makecell{Physical \\Property}
 &\makecell{Function \\Reasoning}
 &\makecell{Nature \\Relation}
 &\makecell{Future \\Prediction}
 &\makecell{Structuralized Image\\-Text Understanding }
 &\makecell{Identity \\Reasoning}
  &\makecell{Social \\Relation }
 &\makecell{Physical \\Relation}
 \\\hline
 \vb Nucleus~\cite{holtzman2019curious}&39.3&68.8&31.3&31.5&20.6&93.2&72.1&17.0	\\
 \gr \cb Nucleus+INTER&\textbf{44.3}&\textbf{77.3}&\textbf{38.0}&\textbf{39.2}&\textbf{24.1}&\textbf{93.8}&\textbf{85.5}&\textbf{22.3}\\
    \vb Greedy~\cite{song2024good}&43.8&82.9&34.6&41.5&27.0&95.5&86.1&25.5\\
\gr \cb Greedy+INTER &\textbf{43.8}&\textbf{82.9}&\textbf{34.6}&\textbf{41.5}&\textbf{27.0}&\textbf{95.5}&\textbf{86.1}&\textbf{25.5}
\\
   \vb Beam~\cite{graves2012sequence, sutskever2014sequence,boulanger2013audio}&43.8&81.6&34.6&41.5&27.0&95.5&86.1&25.5\\
\gr \cb Beam+INTER &\textbf{43.8}&\textbf{82.2}&\textbf{34.6}&\textbf{41.5}&\textbf{27.0}&\textbf{95.5}&\textbf{86.1}&\textbf{25.5}
\\
\vb VCD\textsuperscript{*}~\cite{leng2023mitigating}&41.1&73.7&39.1&40.0&23.1&95.5&83.7&18.1 \\
\gr \cb VCD\textsuperscript{*}+INTER &\textbf{41.1}&\textbf{73.7}&\textbf{39.1}&39.2&\textbf{23.4}&92.6&\textbf{88.4}&\textbf{22.3}\\
\vb 
OPERA\textsuperscript{\dag}~\cite{huang2024opera} &43.8&81.6&34.6&41.5&27.0&95.5&86.1&25.5\\
\gr \cb OPERA\textsuperscript{\dag}+INTER&\textbf{43.8}&\textbf{81.6}&\textbf{34.6}&\textbf{41.5}&\textbf{27.0}&\textbf{95.5}&\textbf{86.1}&\textbf{25.5} \\
\end{tabular}
\caption{\textbf{Evaluating the performance of INTER on MM-Bench~\cite{liu2023mmbench} using LLaVA-v1.5 (7B)~\cite{liu2023improvedllava}}, focusing on attribute reasoning, logic reasoning and relation reasoning subtasks. }
\label{tab:mmbench-llava-l3-r}
\end{table*}
\begin{table*}[ht!]
\tablestyle{2pt}{1.2}
\begin{tabular}{lcccccccccccc}
\hline
 &\multicolumn{5}{c}{Coarse Perception (CP)}  
 &\multicolumn{3}{c}{Cross-instance Fine-grained Perception (FP-C)}
 &\multicolumn{4}{c}{Single-instance Fine-grained Perception (FP-S)}\\ 
 method&\makecell{Image \\Emotion}
 &\makecell{Image \\Topic}
 &\makecell{Image \\Scene}
 &\makecell{Image \\Style}
 &\makecell{Image \\Quality}
 &\makecell{Action \\Recognition}
 &\makecell{Attribute \\Comparision}
 &\makecell{Spatial \\Realtionship}
 &\makecell{Celebrity \\Recognition}
 &\makecell{Object \\Localization}
 &\makecell{Attribute \\Recognition}
 &OCR\\\hline
 \vb Nucleus~\cite{holtzman2019curious} &70.5 & 75.7 & 94.1 & 67.5 & 18.7 & 70.7 & 27.7 & 17.0 & 79.8 & 24.4 & 70.5 &62.8\\
 \gr \cb Nucleus+INTER&\textbf{71.0} & \textbf{77.0} & \textbf{96.1} & \textbf{73.1} & \textbf{19.3} & \textbf{71.2} & \textbf{29.1} & \textbf{15.9} & \textbf{81.1} & \textbf{29.2} & \textbf{73.9} & \textbf{65.4} 

 \\
    \vb Greedy~\cite{song2024good}&\textbf{76.0} & \textbf{76.4} & \textbf{97.1} & \textbf{81.1} & \textbf{22.7} & \textbf{77.7} & \textbf{46.8} & \textbf{25.4} & \textbf{82.8} & \textbf{34.6} & \textbf{73.9} & \textbf{71.8}\\
\gr \cb Greedy+INTER  &\textbf{76.0} & 73.6 & 96.1 & \textbf{83.0} & 12.0 & \textbf{79.5} & 41.8 & 18.6 & 81.8 & 28.6 & \textbf{80.3} & \textbf{71.8}
\\
   \vb Beam~\cite{graves2012sequence, sutskever2014sequence,boulanger2013audio}&76.0 & 76.4 & 97.1& 81.1 & 22.7 & 77.7 & 46.8 & 25.4 & 82.8 & 34.6 & 73.9 & 71.8\\
\gr \cb Beam+INTER  &\textbf{76.0} & \textbf{76.4} & \textbf{97.1} & \textbf{81.1} & \textbf{22.7} & \textbf{77.7} & \textbf{47.1} & \textbf{26.0} & \textbf{82.8} & \textbf{34.7} & \textbf{74.1} & \textbf{71.8}
\\
\vb VCD\textsuperscript{*}~\cite{leng2023mitigating}&71.5 & 79.3 & 96.3 & 75.9 & 17.3 & 72.6 & 32.6 & 17.5 & 79.8 & 26.7 & 75.0 & 66.7 \\
\gr \cb VCD\textsuperscript{*}+INTER
&\textbf{72.5} & \textbf{73.6} & \textbf{96.0} & \textbf{78.8} & \textbf{10.7} & \textbf{74.9} & \textbf{33.3} & \textbf{15.8} & \textbf{81.3} & \textbf{25.2} & \textbf{75.4} & \textbf{68.0} \\
\vb 
OPERA\textsuperscript{\dag}~\cite{huang2024opera} &76.0 &76.4 &97.1 &81.1 &22.7 &77.7&46.8&25.4&82.8&34.6&73.9 &71.8
\\
\gr \cb OPERA\textsuperscript{\dag}+INTER&\textbf{76.0}&\textbf{76.4}&97.0&\textbf{81.1}&\textbf{22.7}&\textbf{78.0}&\textbf{46.8}&\textbf{25.4}&82.7&\textbf{34.6}&\textbf{73.9}&\textbf{71.8} \\
\end{tabular}
\caption{\textbf{Evaluating the performance of INTER on MM-Bench~\cite{liu2023mmbench} using mPLUG-owl2~\cite{ye2024mplug}}, focusing on coarse perception and fine-grained perception subtasks.}
\label{tab:mmbench-mplug-l3-p}
\end{table*}
\begin{table*}[ht!]
\tablestyle{4pt}{1.2}
\begin{tabular}{lccccccccccccccccccccc}
\hline
&
  \multicolumn{3}{c}{Attribute Reasoning (AR)}
 &\multicolumn{3}{c}{Logic Reasoning (LR)}
 &\multicolumn{2}{c}{Relation Reasoning (RR)}\\
 method&\makecell{Physical \\Property}
 &\makecell{Function \\Reasoning}
 &\makecell{Nature \\Relation}
 &\makecell{Future \\Prediction}
 &\makecell{Structuralized Image\\-Text Understanding }
 &\makecell{Identity \\Reasoning}
  &\makecell{Social \\Relation }
 &\makecell{Physical \\Relation}
 \\\hline
 \vb Nucleus~\cite{holtzman2019curious}& 32.4 & 75.3 & 33.0 & 36.2 & 18.8 & 92.1 & 73.3 & 22.3	\\
 \gr \cb Nucleus+INTER& \textbf{33.3} & \textbf{76.6} & \textbf{36.3} & \textbf{40.0}&\textbf{20.9} & \textbf{93.8} & \textbf{79.1} & \textbf{26.9}\\
    \vb Greedy~\cite{song2024good}& 33.3 & 82.2 & 36.9 & 45.4 & 23.1 & 93.8 & 88.4 & 38.3
\\
\gr \cb Greedy+INTER & \textbf{36.1} & 79.1 & \textbf{41.3} & \textbf{46.9} & 21.3 & \textbf{96.0} & 86.1 & 25.5

\\
   \vb Beam~\cite{graves2012sequence, sutskever2014sequence,boulanger2013audio}& 33.3 & 82.2 & 36.9 & 45.4 & 24.5 & 93.8& 88.4 & 34.0
\\
\gr \cb Beam+INTER & \textbf{33.9} & \textbf{82.2} & \textbf{36.9} & \textbf{45.4} & \textbf{24.5} & \textbf{93.8} & \textbf{88.4} & \textbf{38.2}

\\
\vb VCD\textsuperscript{*}~\cite{leng2023mitigating}& 32.0 & 75.3 & 37.4 & 36.9 & 20.9 & 92.6 & 77.3 & 25.5 \\
\gr \cb VCD\textsuperscript{*}+INTER & \textbf{34.3} & \textbf{76.0} & \textbf{39.3} & \textbf{43.9} & \textbf{20.2} & \textbf{95.5} & \textbf{79.1} & \textbf{20.2}\\
\vb 
OPERA\textsuperscript{\dag}~\cite{huang2024opera}&\textbf{33.3} &\textbf{82.2} &\textbf{36.9} &\textbf{45.4} &\textbf{23.1} &\textbf{93.8} &\textbf{88.4} &\textbf{34.0}\\
\gr \cb OPERA\textsuperscript{\dag}+INTER&\textbf{33.8}&82.1&\textbf{36.9}&\textbf{45.4}&\textbf{24.5}&\textbf{93.8}&\textbf{88.9}&\textbf{38.3} \\
\end{tabular}
\caption{\textbf{Evaluating the performance of INTER on MM-Bench~\cite{liu2023mmbench} using mPLUG-owl2~\cite{ye2024mplug}}, focusing on attribute reasoning, logic reasoning and relation reasoning subtasks. }

\label{tab:mmbench-mplug-l3-r}
\end{table*}
\begin{table*}[h!]
\tablestyle{2pt}{1.2}
\begin{tabular}{lcccccccccccc}
\hline
 &\multicolumn{3}{c}{Coarse Perception (CP)}  
 &\multicolumn{3}{c}{Fine-grained Perception (FP)}
 &\multicolumn{3}{c}{Instance Reasoning (IR)}\\ 
 method&\makecell{Image Scene \\\& Topic}
 &\makecell{Image Style \\\& Quality}
 &\makecell{Image \\Emotion}
 &\makecell{Object \\Counting}
 &\makecell{Recognition}
 &\makecell{Localization}
 &\makecell{Single-Instance \\Reasoning}
 &\makecell{Cross-Instance \\Attribute Reasoning}
 &\makecell{Cross-Instance \\Relation Reasoning}\\\hline
 \vb Nucleus~\cite{holtzman2019curious}&45.4&66.7&51.6&17.4&25.4&25.0&56.6&29.2&21.0 \\
 \gr \cb Nucleus+INTER&\textbf{48.2} & \textbf{73.1} & \textbf{58.1} & \textbf{25.0} & \textbf{37.3} & \textbf{27.5} & 51.5 & \textbf{33.7} & \textbf{32.3} \\
    \vb Greedy~\cite{song2024good}&48.9&
74.4&
67.7&
20.7&
29.7&
20.0&
52.5&
31.5&
32.3\\
\gr \cb Greedy+INTER & 
48.2&
66.7&
58.1&
\textbf{32.6}&
27.1&
\textbf{27.5}&
\textbf{56.6}&
\textbf{34.8}&
\textbf{41.9}\\
   \vb Beam~\cite{graves2012sequence, sutskever2014sequence,boulanger2013audio}&48.9&71.8&67.7&21.7&25.4&17.5&51.5&28.1&40.3\\
\gr \cb Beam+INTER &47.5 & 69.2 & \textbf{67.7} & \textbf{26.1} & \textbf{27.1} & \textbf{27.5} & \textbf{56.6} & \textbf{28.1} & \textbf{41.9}\\
\vb VCD\textsuperscript{*}~\cite{leng2023mitigating}&44.0&70.5&61.3&27.2&22.9&22.5&48.5&32.6&29.0\\
\gr \cb VCD\textsuperscript{*}+INTER&\textbf{47.5} & \textbf{70.5} & 54.8 & 26.1 & \textbf{28.8} & \textbf{22.5} & \textbf{52.5} & 31.5 & \textbf{32.3}
\\
\vb 
OPERA\textsuperscript{\dag}~\cite{huang2024opera} &49.6&73.1&67.7&22.8&26.3&17.5&52.5&31.5&35.5\\
\gr \cb OPERA\textsuperscript{\dag}+INTER &45.4&71.8&\textbf{67.7}&\textbf{31.5}&\textbf{31.4}&\textbf{20.0}&\textbf{56.6}&\textbf{31.5}&\textbf{37.1}\\
\end{tabular}
\caption{\textbf{Evaluating the performance of INTER on MMStar~\cite{chen2024we} using LLaVA-v1.5 (7B)~\cite{liu2023improvedllava},} focusing on coarse perception, fine-grained perception and instance reasoning substasks. }
\label{tab:MMStar-llava-l3-p}
\end{table*}
\begin{table*}[h!]
\tablestyle{1pt}{1.2}
\begin{tabular}{lcccccccccccc}
\hline
 &\multicolumn{3}{c}{Logit Reasoning (LR)}  
 &\multicolumn{3}{c}{Science and Technology (ST)}
 &\multicolumn{3}{c}{Math (MA)}\\ 
 method&\makecell{Code Sequence\\Reasoning}
 &\makecell{Diagram \\Reasoning}
 &\makecell{Common \\Reasoning}
 &\makecell{Biology \\\&Chemistry \\\& Physics}
 &\makecell{Electronics \\\& Energy \\\& Mechanical eng.}
 &\makecell{Geography \\\& Earth Science \\\& Agriculture}
 &\makecell{Geometry}
 &\makecell{Numeric Commonsense\\\& Calculation}
 &\makecell{Statistical \\Reasoning}\\\hline
 \vb Nucleus~\cite{holtzman2019curious}&23.1&19.1&26.7&16.7&20.5&20.7&19.8&33.3&27.7 \\
 \gr \cb Nucleus+INTER&\textbf{25.6} & \textbf{19.1} & \textbf{34.7} & \textbf{16.0} & 17.9 & \textbf{20.7} & \textbf{30.2} &31.3 & 19.3
\\
   \vb Greedy~\cite{song2024good}
&23.1&
22.7&
33.7&
12.5&
15.4&
17.2&
25.6&
27.1&
16.9\\
\gr \cb Greedy+INTER &
\textbf{23.1}&
21.8&
\textbf{39.6}&
\textbf{15.3}&
10.3&
\textbf{20.7}&
17.4&
18.8&
\textbf{24.1}
\\
   \vb Beam~\cite{graves2012sequence, sutskever2014sequence,boulanger2013audio}&33.3&24.5&31.7&13.2&12.8&22.4&25.6&31.3&19.3\\
\gr \cb Beam+INTER &\textbf{35.9} & 21.8 & \textbf{32.7} & \textbf{16.0} & \textbf{25.6} & \textbf{27.6} & 24.4 & 27.1 & \textbf{22.9}
\\
\vb VCD\textsuperscript{*}~\cite{leng2023mitigating}&20.5&20.0&34.7&18.1&15.4&17.2&27.9&20.8&26.5\\
\gr \cb VCD\textsuperscript{*}+INTER&\textbf{23.1} & \textbf{23.6} & \textbf{36.6} & 17.4 &7.7 & \textbf{17.2} & 17.4& \textbf{27.1} & \textbf{27.7}

\\
\vb 
OPERA\textsuperscript{\dag}~\cite{huang2024opera} &33.3&24.5&31.7&13.2&12.8&22.4&25.6&31.3&19.3\\
\gr \cb OPERA\textsuperscript{\dag}+INTER &\textbf{35.9}&21.8&\textbf{31.7}&\textbf{15.3}&\textbf{25.6}&\textbf{27.6}&23.3&27.1&\textbf{22.9}\\
\end{tabular}
\caption{\textbf{Evaluating the performance of INTER on MMStar~\cite{chen2024we} using LLaVA-v1.5 (7B)~\cite{liu2023improvedllava},} focusing on logit reasoning, science technology and math capability substasks.}
\label{tab:MMStar-llava-l3-2}
\end{table*}
\begin{table*}[h!]
\tablestyle{2pt}{1.2}
\begin{tabular}{lcccccccccccc}
\hline
 &\multicolumn{3}{c}{Coarse Perception (CP)}  
 &\multicolumn{3}{c}{Fine-grained Perception (FP)}
 &\multicolumn{3}{c}{Instance Reasoning (IR)}\\ 
 method&\makecell{Image Scene \\\& Topic}
 &\makecell{Image Style \\\& Quality}
 &\makecell{Image \\Emotion}
 &\makecell{Object \\Counting}
 &\makecell{Recognition}
 &\makecell{Localization}
 &\makecell{Single-Instance \\Reasoning}
 &\makecell{Cross-Instance \\Attribute Reasoning}
 &\makecell{Cross-Instance \\Relation Reasoning}\\\hline
 \vb Nucleus~\cite{holtzman2019curious}&43.9 & 56.4 & 64.5 & 27.2 & 24.6 & 20.0 & 49.5 & 28.1 & 41.9  

 \\
 \gr \cb Nucleus+INTER&\textbf{47.5} & \textbf{58.9} & \textbf{70.9} & \textbf{31.5} & \textbf{27.9} & 15.0 & \textbf{53.5} & \textbf{31.5} & 30.6 
\\
\vb Greedy~\cite{song2024good} &46.1&
59.0&
71.0 &
26.1&
30.5&
12.5 & 
53.5&
30.3&
33.9& \\
\gr \cb Greedy+INTER &\textbf{48.9}&
\textbf{61.5}&
67.7 & \textbf{33.7}&
29.7&
\textbf{20.0} & \textbf{55.6}&
\textbf{32.6}&
\textbf{35.5} \\
\vb Beam~\cite{graves2012sequence, sutskever2014sequence,boulanger2013audio} &45.4 & 57.7 & 70.9 & 27.2 & 29.7 & 12.5 & 53.5 & 29.2 & 41.9 \\
\gr \cb Beam+INTER &\textbf{46.1} & \textbf{57.7} & \textbf{70.9} & \textbf{28.2} & 26.3 & 10.0 & \textbf{53.5} & 28.1 & 32.3 \\
\vb VCD\textsuperscript{*}~\cite{leng2023mitigating}&46.8 & 58.9 & 64.5 & 27.2 & 28.2 & 20.0 &  46.5 & 28.1 & 30.6 
\\
\gr \cb VCD\textsuperscript{*}+INTER
&\textbf{47.5} &56.4 & \textbf{67.7} & \textbf{28.3} & 24.6 & \textbf{22.5} & \textbf{51.5} & \textbf{30.3} & \textbf{33.9}

\\
\vb 
OPERA\textsuperscript{\dag}~\cite{huang2024opera} &45.4 & 58.9 & 70.9 &  27.2 & 29.7 & 15.0 &  53.5 & 29.2 & 40.3 
\\
\gr \cb OPERA\textsuperscript{\dag}+INTER &\textbf{46.1} & 57.7 & 70.9 &  \textbf{28.3} & 26.3 & 10.0 & \textbf{53.5} & 28.1 & 33.9 
\\
\end{tabular}
\caption{\textbf{Evaluating the performance of INTER on MMStar~\cite{chen2024we} using mPLUG-owl2~\cite{ye2024mplug},} focusing on coarse perception, fine-grained perception and instance reasoning substasks. }
\label{tab:MMStar-mplug-l3-1}
\end{table*}
\begin{table*}[ht!]
\tablestyle{1pt}{1.2}
\begin{tabular}{lcccccccccccc}
\hline
 &\multicolumn{3}{c}{Logit Reasoning (LR)}  
 &\multicolumn{3}{c}{Science and Technology (ST)}
 &\multicolumn{3}{c}{Math (MA)}\\ 
 method&\makecell{Code Sequence\\Reasoning}
 &\makecell{Diagram \\Reasoning}
 &\makecell{Common \\Reasoning}
 &\makecell{Biology \\\&Chemistry \\\& Physics}
 &\makecell{Electronics \\\& Energy \\\& Mechanical eng.}
 &\makecell{Geography \\\& Earth Science \\\& Agriculture}
 &\makecell{Geometry}
 &\makecell{Numeric Commonsense\\\& Calculation}
 &\makecell{Statistical \\Reasoning}\\\hline
 \vb Nucleus~\cite{holtzman2019curious}&25.6& 21.8 & 34.7 & 15.1 & 21.7 & 20.7  & 22.4 & 22.9 & 22.1\\
 \gr \cb Nucleus+INTER&12.8& 18.2 & \textbf{42.6} & \textbf{19.2} & \textbf{36.9} & \textbf{24.1}  & \textbf{19.8} & \textbf{20.8} & \textbf{26.7}
\\
   \vb Greedy~\cite{song2024good}& 30.8&
24.6&
32.7 & 11.0&
8.7&
17.2 & 18.1
&25.0&
19.8\\
\gr \cb Greedy+INTER & 20.5&
\textbf{24.6}&
\textbf{41.6 }& \textbf{15.8}&
\textbf{23.9}&
15.5& \textbf{19.0}&
\textbf{27.1}&
\textbf{31.4}
\\
   \vb Beam~\cite{graves2012sequence, sutskever2014sequence,boulanger2013audio}&  25.6 & 25.5 & 33.7 & 11.6 & 6.5 & 17.2  & 18.9 & 25.0 & 27.9\\
\gr \cb Beam+INTER & \textbf{25.6} & 24.5 & \textbf{36.6} & \textbf{12.3} & \textbf{15.2} & \textbf{17.2} &  \textbf{19.8} & 22.9 & 26.7
\\
\vb VCD\textsuperscript{*}~\cite{leng2023mitigating}& 28.2 & 20.9 & 29.7 & 23.3 & 10.9 & 13.8 &  23.3 & 33.7 & 33.7\\
\gr \cb VCD\textsuperscript{*}+INTER & 17.9 & \textbf{25.5} & \textbf{44.6} & 15.8 & \textbf{21.7} & \textbf{17.2} & \textbf{25.0} & 27.1 & 29.1

\\
\vb 
OPERA\textsuperscript{\dag}~\cite{huang2024opera}&  25.6 & 23.6 & 34.7 &11.6 & 6.5 & 17.2 &  19.0 & 25.0 & 24.4\\
\gr \cb OPERA\textsuperscript{\dag}+INTER & \textbf{25.6} & \textbf{25.5} & \textbf{35.6} & \textbf{13.0} & \textbf{17.4} & \textbf{17.2} & \textbf{21.6} & \textbf{25.0} & \textbf{26.7}\\
\end{tabular}
\caption{\textbf{Evaluating the performance of INTER on MMStar~\cite{chen2024we} using mPLUG-owl2~\cite{ye2024mplug},} focusing on logit reasoning, science, technology, and math capability subtasks.}
\label{tab:MMStar-mplug-l3-2}
\end{table*}

\end{document}